**Driving Accurate Allergen Prediction with Protein Language Models and Generalization-Focused Evaluation**

Brian Shing-Hei Wong[1], Joshua Mincheol Kim[1], Sin-Hang Fung[1,2], Qing Xiong[3], Kelvin Fu-Kiu Ao[1], Junkang Wei[1], Ran Wang[1], Dan Michelle Wang[1], Jingying Zhou[1], Bo Feng[1], Alfred Sze-Lok Cheng[1], Kevin Y. Yip[4,5,6,7,*], Stephen Kwok-Wing Tsui[1,8,*], Qin Cao[1,2,8,*]

[1] School of Biomedical Sciences, The Chinese University of Hong Kong, Shatin, New Territories, Hong Kong SAR, China

[2] Shenzhen Research Institute, The Chinese University of Hong Kong, Shenzhen, China

[3] Department of Health Technology and Informatics, The Hong Kong Polytechnic University, Hung Hom, Kowloon, Hong Kong SAR, China

[4] Center for Data Sciences, Sanford Burnham Prebys Medical Discovery Institute, La Jolla, CA, USA

[5] Cancer Genome and Epigenetics Program, NCI-Designated Cancer Center, Sanford Burnham Prebys Medical Discovery Institute, La Jolla, CA, USA

[6] Center for Neurologic Diseases, Sanford Burnham Prebys Medical Discovery Institute, La Jolla, CA, USA

[7] Department of Computer Science and Engineering, The Chinese University of Hong Kong, Shatin, New Territories, Hong Kong SAR, China

[8] Hong Kong Bioinformatics Centre, The Chinese University of Hong Kong, Shatin, New Territories, Hong Kong SAR, China

[*] Correspondence

To whom correspondence should be addressed. Emails: kyip@sbpdiscovery.org (KYY), kwtsui@cuhk.edu.hk (SKWT), qcao@cuhk.edu.hk (QC)

**Abstract**

Allergens, typically proteins capable of triggering adverse immune responses, represent a significant public health challenge. To accurately identify allergen proteins, we introduce Applm (Allergen Prediction with Protein Language Models), a computational framework that leverages the 100-billion parameter xTrimoPGLM protein language model. We show that Applm consistently outperforms seven state-of-the-art methods in a diverse set of tasks that closely resemble difficult real-world scenarios. These include identifying novel allergens that lack similar examples in the training set, differentiating between allergens and non-allergens among homologs with high sequence similarity, and assessing functional consequences of mutations that create few changes to the protein sequences. Our analysis confirms that xTrimoPGLM, originally trained on one trillion tokens to capture general protein sequence characteristics, is crucial for Applm's performance by detecting important differences among protein sequences. In addition to providing Applm as open-source software, we also provide our carefully curated benchmark datasets to facilitate future research.

## Introduction

Allergens are typically proteins that trigger abnormal immune responses in susceptible individuals. These responses occur when the immune system inappropriately identifies substances that are typically well-tolerated by most individuals as threats, leading to an exaggerated reaction and the excessive release of mediators such as histamine[1,2]. Resulting symptoms range from mild effects, such as skin rashes, sneezing, and watery eyes, to severe and life-threatening reactions like anaphylaxis[1]. Collectively, these symptoms and their underlying dysfunction define allergy, a persistent global health challenge. Recent epidemiological estimates suggest that 30%-40% of the global population is affected by some form of allergy[3-7]. Consequently, accurate allergen identification is critical for mitigating health risks and guiding targeted interventions to reduce the burden on healthcare systems.

Clinical diagnostic methods such as serum-specific IgE testing[8] and skin prick tests[9], which determine if an individual is sensitized, and gold standard tests like double-blind placebo-controlled food challenges[9,10], which assess whether the individual exhibits an allergic response (i.e., is clinically sensitive) to a specific substance, are crucial in allergy diagnosis. However, it is difficult for these individual-focused methods to reveal which specific protein within the substance is responsible, and whether the response is representative at the population level, which typically requires evidence from a clinically significant number of sensitized individuals. To identify allergens, proteins in the allergenic substance are extracted and screened for IgE reactivity using the sera of sensitized individuals. According to major allergen databases such as World Health Organization/International Union of Immunological Societies (WHO/IUIS) Allergen Nomenclature database[11], the COMprehensive Protein Allergen REsource (COMPARE) database[12], and the AllergenOnline database[13], a protein is typically defined as an allergen if it binds IgE from a clinically significant number of sensitized individuals, is molecularly characterized, and, in most cases, is further validated by additional immunological assays. While accurate allergen identification supports clinical diagnostics by enabling more targeted testing, and a comprehensive allergen catalog is vital for advancing biological understanding and therapeutic development, this entire experimental identification process is time-consuming, resource-intensive, particularly considering the sample requirements, and difficult to scale. As the diversity of novel proteins continues to expand, particularly through advances in food production, genetic engineering, and synthetic biology, scalable and reliable computational methods are urgently needed to achieve more efficient and cost-effective allergen identification.

Early computational methods relied on rule-based strategies. For example, the Food and Agriculture Organization (FAO) and WHO guidelines classify proteins as potential allergens based on sequence identity thresholds to known allergens[14]. While simple, these methods tend to be overly sensitive, often resulting in high false-positive rates[15]. Consequently, significant resources are often wasted on experimentally verifying numerous non-allergenic proteins, potentially also hindering the development or use of safe and beneficial novel proteins. This limitation has driven the development of more accurate models.

In recent years, various machine learning (ML) methods have been proposed for allergen prediction (Supplementary Table 1). These methods[16-32] typically employ classifiers such as k-nearest neighbors (k-NN), random forests (RF), and deep learning, using protein sequence features like amino acid composition, physicochemical descriptors, and motifs as input. While effective to some extent, such features may fail to capture the full biological complexity of protein sequences. Meanwhile, protein language models (pLMs), pre-trained on vast and diverse protein sequences, have emerged as powerful tools for learning rich, contextualized sequence representations (i.e., embeddings) that can better reflect these complexities. Models such as the ESM[33] and ProtTrans[34] families have achieved state-of-the-art results across diverse protein-related tasks, including structure prediction, function annotation, and mutational effect analysis. Recently, three studies have explored the application of pLM embeddings on allergen prediction[25,31,32], reporting state-of-the-art performance and highlighting the utility of pLMs in this domain.

Most existing ML methods for allergen prediction, including these three latest pLM-based methods, have been reported to achieve very good performance, with the area under the receiver operating characteristic curve (AUROC) often exceeding 0.9. However, as our analysis in this study reveals, a critical disconnect exists: high reported performance in allergen prediction, often achieved under idealized evaluation settings, frequently fails to translate to robust real-world efficacy. This gap largely stems from evaluation methodologies that oversimplify the predictive challenge, notably by inadequately addressing sequence similarity complexities. Consequently, models often underperform when facing the true intricacies of practical allergen prediction. The specific shortcomings of current evaluation practices that contribute to this performance discrepancy will be dissected below.

A primary concern is high sequence similarity between training and test sets (hereafter referred to as “inter-split” similarity). Ensuring sufficient dissimilarity between these sets is a fundamental principle for reliable model evaluation. While this principle is widely accepted in theory in many protein prediction tasks[35-38], its practical implementation often falls short. In allergen prediction, however, the problem is more fundamental, as the principle itself is frequently overlooked. Such high inter-split similarity can lead to overly optimistic performance estimates[39], as models may merely memorize training data rather than learn generalizable allergenic determinants. This issue is particularly critical in allergen prediction for two reasons. First, newly identified allergens can significantly differ from known ones[40-42]. As a result, if a model is mostly tested on sequences highly similar to the training sequences, a high test performance could give a wrong message that the model can accurately predict these new sequences. Second, the inflated test performance stemming from such high inter-split similarity can mask the model’s potential insensitivity to subtle yet critical sequence differences. Consequently, the model is likely to predict sequences highly similar to allergens in the training set all as allergens, failing to distinguish cases where highly similar homologous sequences differ in allergenicity or where minor mutations alter allergenicity.

Our examination of current evaluation practices, supported by our survey of 17 published allergen prediction methods, indicates prevalent shortcomings (Supplementary Table 1). 15 of 17 relied on internal cross-validation (CV) and all the 15 studies inadequately controlled inter-split similarity: 12 entirely disregarded it, while the remaining 3 employed clustering tools like CD-HIT[43,44], insufficient for proper similarity-based partitioning. The inadequacy of CD-HIT for ensuring desired similarity separation has also been recently revealed in other protein prediction contexts[35,45-48] (Supplementary text, Supplementary Fig. 1).

Furthermore, distinct from high inter-split similarity that can inflate performance via memorization, low sequence similarity between allergens and non-allergens (hereafter referred to as "inter-class" similarity) can also inflate test performance by making classification artificially easy. Such simplification in training simultaneously hinders models from learning to handle difficult real-world scenarios, especially the subtle distinctions discussed earlier, such as highly similar homologs or allergenicity changes from a few mutations. Despite its importance, our survey of 17 studies shows that only 2 appropriately controlled for inter-class similarity, while the remaining 15 either omitted this control entirely (10 studies) or implemented it counterproductively by enforcing dissimilarity (5 studies), which further simplified the classification task.

To address these critical deficiencies in current evaluation practices and to establish a more robust framework for allergen prediction, in this work, we introduce Applm (Allergen Prediction with Protein Language Models), a novel framework that consistently outperforms existing methods for allergen prediction. Applm advances this field through several key contributions. First, it leverages state-of-the-art pLMs, including the first application of the 100-billion (100B) parameter xTrimoPGLM[49] in allergen prediction, to capture richer sequence representations. Second, and critically, our work pioneers the comprehensive model evaluation on multiple diverse real-world benchmark datasets, each reflecting distinct challenges critical to assessing true model generalization. Third, we developed a novel similarity-aware evaluation pipeline that addresses critical similarity biases, enabling internal CV to better mimic real-world challenges. Finally, our study investigates other key factors influencing predictive performance, offering actionable insights for future model refinement.

Our work not only establishes a new pLM-driven state-of-the-art in allergen prediction but also introduces an essential real-world benchmark to foster and validate continued research in this domain. Moreover, our similarity-aware evaluation pipeline is broadly applicable beyond allergen prediction and can benefit a wide range of other protein classification tasks susceptible to similar evaluation pitfalls.

## Overview of Applm

Applm's core design focuses on enhancing allergen prediction accuracy while ensuring the reliability of its performance assessment, a particularly critical aspect when dealing with challenging real-world scenarios such as identifying novel allergens and discriminating subtle, allergenicity-critical sequence variations (Methods and Fig. 1).

A cornerstone of Applm is its strategic utilization of multiple state-of-the-art pLMs. Its main approach, a two-stage process, first utilizes four pre-trained pLMs: xTrimoPGLM-100B, its lightweight version xTrimoPGLM-10B, ESM-2[33], and ProtTrans T5 (ProtT5)[34], to transform protein sequences into rich numerical representations (i.e., frozen embeddings). These representations are then fed into an RF classifier to make the final allergenicity prediction. Pioneering the application of the massive-scale 100B parameter xTrimoPGLM in allergen prediction, we anticipate that its extensive scale offers the potential to capture more nuanced and complex sequence determinants of allergenicity than previously utilized, smaller models. Unless specified otherwise, "Applm" hereafter refers to this main configuration. For comparison, an exploratory fine-tuned variant was also assessed for its potential (Methods). Critically, the robustness and practical utility of Applm are rigorously ascertained through comprehensive evaluations on multiple diverse real-world benchmark datasets, each curated to reflect distinct and challenging generalization scenarios previously underexplored in this field. This stringent external validation is further complemented by our novel similarity-aware evaluation pipeline. This pipeline systematically controls for both inter-split (between training and test sets) and inter-class (between allergens and non-allergens) sequence similarities, ensuring the integrity of model development and evaluation by making internal CV a reliable and flexible mimic of real-world challenges. Such comprehensive control mitigates the overly optimistic performance often reported in previous studies.

As a result of these innovations, Applm demonstrates superior predictive performance that has been robustly validated, outperforming existing methods across benchmark datasets.

### Applm outperforms existing methods on rigorous real-world allergen prediction benchmark datasets

To rigorously assess Applm's performance and generalization capabilities, we benchmarked it against seven leading allergen prediction methods, including two pLM-based methods (Methods). This evaluation was conducted on a carefully curated suite of six external test sets, grouped into three diverse and challenging real-world scenarios where accurate allergen prediction is critical (Methods and Supplementary text). The first scenario, "By Date", tests the performance in identifying newly discovered allergens using a temporal cutoff. The second, "Homologs", assesses the performance in distinguishing between allergenic and non-allergenic proteins within highly homologous families. The third, "Mutations", evaluates the performance in predicting how minor sequence variations alter a protein's allergenicity.

Across the comprehensive benchmark suite, Applm demonstrated clear superiority over all seven competing methods (Fig. 2). This leading performance was validated across multiple metrics, including average AUROC (Fig. 2a), the area under the precision-recall curve (AUPRC; Fig. 2b), and statistical rank-based analysis via Friedman and Nemenyi post-hoc tests (Fig. 2c and d). Quantitatively, Applm achieved an average AUROC of 0.872, exceeding competitors by a margin of 0.065–0.37. Similarly, its average AUPRC of 0.700 dramatically surpassed the random baseline (0.202) and outperformed other methods by 0.104–0.500. This dominance was remarkably consistent across benchmark datasets: Applm ranked first in 14 of 18 test settings by AUROC and in all 18 settings by AUPRC (Supplementary Fig. 2 and Supplementary Fig. 3). In the few remaining cases, it secured a top-four position at worst. Such consistent, top-tier performance across diverse challenges underscores its reliability and effectiveness.

Among the four pLMs integrated into Applm, xTrimoPGLM-100B achieved the highest overall performance (Fig. 2a and b), likely due to its larger parameter size and embedding size. However, the differences were not statistically significant, suggesting that Applm is robust to the choice of pLM backbone. Further statistical tests confirmed that Applm with the xTrimoPGLM-100B backbone significantly outperformed six of the seven competing methods and maintained a slight but consistent edge over the remaining AlgPred2 (Fig. 2c and d).

Among the seven methods evaluated, two deep learning models, pLM4Alg and Alg-MFDL, also utilized frozen embeddings from ESM-2 and ProtT5. However, despite extensive hyperparameter tuning (Methods), neither outperformed Applm (Fig. 2). A possible explanation is that RF can more readily leverage high-dimensional embeddings without extensive tuning, and may generalize better in settings with limited training data[50,51]. In contrast, deep neural networks often require more data and careful hyperparameter tuning to perform well[51,52]. To further investigate this, we replaced Applm's RF with a feed-forward neural network (FFNN; Methods). The FFNN alternative indeed showed a notable performance drop (Supplementary Fig. 4).

Interestingly, the fine-tuned variants did not improve upon Applm using frozen embeddings (Fig. 2). This may be due to the relatively limited size of the training sets compared to the scale of pre-training, which limits the effectiveness of fine-tuning[53,54]. In addition, previous studies have shown that fine-tuning may degrade performance under distributional shift by distorting pre-trained representations[55,56]. Given the diversity and complexity of these real-world test sets, using frozen embeddings with a simple, stable classifier may offer better performance.

While Applm's overall performance was strong, it varied across the diverse external benchmark datasets. To understand this variation, we investigated two key factors: the intrinsic difficulty of each benchmark and its generalization demand. We used average inter-class similarity as a proxy for intrinsic difficulty (where higher values mean harder tasks) and average inter-split similarity as a proxy for generalization demand (where lower values mean a greater generalization challenge; Methods and Supplementary Fig. 5). Our analysis revealed that intrinsic difficulty was the predominant factor in driving performance. We found a strong, statistically significant negative correlation between Applm's AUROC and the average inter-class similarity (both Pearson correlation coefficient and Spearman's rank correlation coefficient $< -0.9$). In contrast, we found no significant correlation between performance and the average inter-split similarity, likely overshadowed by the dominant influence of intrinsic difficulty. This suggests that while generalization is an inherent challenge, for the benchmark datasets under study, their intrinsic difficulty emerged as the more dominant factor influencing performance.

Overall, these results demonstrate that Applm is a robust and effective approach for allergen prediction, setting a new state-of-the-art across diverse real-world scenarios. The profound influence of sequence similarity on prediction difficulty, as observed, underscores the importance of carefully similarity-controlled evaluation, a topic we will explore in the next section.

## pLM embeddings used by Applm outperform conventional encodings under similarity-aware evaluation

Following Applm's state-of-the-art performance on diverse external benchmark datasets, we aimed to dissect the contribution of its core component, the pLM embeddings, by comparing them to conventional encodings used by most previous studies. While external benchmark datasets are vital for assessing overall real-world efficacy, this particular ablation study benefits from a more controlled evaluation by systematically creating a spectrum of inter-split and inter-class similarities. Our similarity-aware internal CV pipeline is designed for this purpose, explicitly addressing the previously discussed limitation of tools like CD-HIT in guaranteeing desired similarity separation for rigorous benchmarking. This refined approach, with its precise controls, thus offers a robust and standardized platform for comprehensively evaluating performance under varying sequence similarity conditions. We will first outline this pipeline's key principles before presenting the comparative results (Fig. 3a).

Our similarity-aware pipeline partitions protein sequences into training and test splits based on pre-defined inter-split and inter-class thresholds (denoted as $T_S$ and $T_C$, respectively; Methods). The main steps involve clustering similar positive sequences into distinct splits, removing similarity violations among positives across splits (i.e., positives that would cause inter-split similarity to exceed $T_S$), pairing negatives with positives based on $T_C$, and eliminating similarity violations among negatives across splits (i.e., negatives that would cause inter-split similarity to exceed $T_S$). Using our pipeline, we partitioned the dataset into three splits for internal CV. We assessed sequence similarity across splits and confirmed that no similarity violations occurred, ensuring reliable evaluation of model performance (Fig. 3b and Supplementary Fig. 6).

Across all inter-split and inter-class similarity thresholds, models equipped with pLM embeddings consistently and substantially outperformed those using conventional encodings, namely One-Hot Encoding (OHE) or BLOSUM62 (BL62)[57] (Fig. 4a and Supplementary Fig. 7a). Quantitatively, models using pLM embeddings achieved an overall average AUROC between 0.844 and 0.864, standing in stark contrast to the 0.742 and 0.684 achieved by models using OHE and BL62, respectively (Fig. 4b). A similarly large performance gap was evident for the AUPRC metric (Supplementary Fig. 7b). These results suggest that the superior performance of Applm is primarily driven by the representational capacity of pLM embeddings.

We next evaluated the impact of inter-split and inter-class sequence similarities on model performance. Specifically, we hypothesized that performance would be higher in "easy" scenarios, defined by high inter-split similarity thresholds ($T_S$) and low inter-class similarity thresholds ($T_C$), and lower in "difficult" ones (i.e., low $T_S$ and high $T_C$). AUROC was used as the primary metric, as AUPRC is difficult to compare across test sets with varying class imbalance. Confirming our hypothesis, Applm's performance mapped directly to the scenario difficulty (Fig. 4c). It achieved a markedly high average AUROC of 0.942 in the easiest scenario and its lowest of 0.722 in the most difficult one. The heatmap vividly illustrates this trend, revealing a distinct performance gradient from the top-left corner to the bottom-right one. This gradient is quantifiable along both axes: when averaging over all $T_C$ values, the average AUROC declined from 0.884 to 0.802 with decreasing $T_S$; conversely, when averaging over all $T_S$ values, it dropped from 0.944 to 0.777 with increasing $T_C$. A more granular analysis of the heatmap reinforces these findings. Within each row (representing a fixed $T_S$), model performance generally declined as $T_C$ increased. The column-wise analysis (representing a fixed $T_C$) reveals a more nuanced pattern. Specifically, in the easy scenario where $T_C$ = 0, performance remained exceptionally high and stable across all $T_S$ values. For all other $T_C$ levels, performance showed a strong downward trend as $T_S$ decreased. Overall, these observations indicate that model performance is influenced jointly and systematically by inter-split and inter-class similarities.

To ensure our findings were not an artifact of varying training set sizes (as stricter thresholds reduce instance counts), we performed a control experiment. We subsampled all datasets to an identical size (the "Minimal" strategy in Methods) and repeated the analysis. The results confirm the robustness of our observations, with all previously identified performance trends remaining highly consistent under the Minimal strategy (Supplementary Fig. 8 and Supplementary Fig. 9).

Together, these findings underscore the necessity of evaluating models across a spectrum of inter-split and inter-class similarities. Such a multi-faceted evaluation is critical for truly assessing generalization, as this similarity variation is inherent in real-world datasets. Notably, pLM embeddings consistently demonstrate superior performance across all similarity-aware conditions, underscoring their robustness and representational strength.

**Intrinsic difficulty, sequence length distribution, training set size, and training set imbalance impact model performance**

In addition to the superior expressive power of pLM embeddings, we next systematically analyzed other key factors influencing model performance, aiming to uncover actionable insights for advancing allergen prediction models.

A key question in predictive modeling is how to tailor a training set for a specific inference task. We investigated the principle of "difficulty matching," using inter-class similarity as a proxy for task intrinsic difficulty. Our hypothesis was that model performance would be optimal when the difficulty of the training set aligns with that of the test set. In our controlled internal CV (Methods), this hypothesis was clearly supported (Fig. 5a). For a fixed test set difficulty, model performance was generally highest when the training set's $T_C$ value matched that of the test set. A minor exception occurred in the most difficult scenario (test set $T_C$ = 0.7), where a training set with $T_C$ = 0.6 achieved a marginally better result than the matched set. This trend also held true when controlling for training set size using our Minimal strategy (Supplementary Fig. 10). These findings suggest that neither using an unfiltered training set nor deliberately constructing a highly difficult training set leads to universally optimal results. Instead, aligning the difficulty levels of the training and test sets appears to be the most effective strategy. To assess the generalization of the "difficulty matching" principle, we extended our analysis to the external benchmark. This comparison was fully feasible for the Tropomyosin external test set and partially for Arginine Kinase (Methods). Notably, the performance trend for Tropomyosin visually supported our hypothesis, peaking when the training and test set difficulties were matched, although the differences were not statistically significant (Supplementary Fig. 11). This suggests that in real-world applications, the principle's effects are likely modulated by a variety of factors not fully accounted for by our $T_C$ proxy. This highlights the need for more comprehensive training strategies and better-characterized external benchmarks. Nevertheless, the principle remains a valuable conceptual guide. In practice, since test labels are unavailable, task difficulty can be approximated using prior biological knowledge (e.g., assessing if test proteins belong to the same family) or by measuring intra-set sequence similarity.

We next investigated how differences in the sequence length distribution between allergens and non-allergens impact model performance. While sequence length itself may have limited biomedical relevance to allergenicity, a systematic difference, such as allergens being generally shorter than non-allergens in protein databases (Supplementary Fig. 12a), can create a "shortcut." This can allow the model to rely on length as a simplistic heuristic to classify proteins (Supplementary Fig. 13), rather than learning biologically meaningful patterns from the amino acid sequence. To address this potential bias, we adopted a "Length Control" strategy, proposed by previous work on antimicrobial peptide prediction[58], which explicitly matches the length distributions of positive and negative sequences in the training set (Methods). We then compared the performance of Length Control against our main strategy (i.e., "Hard Balance") on our external benchmark. According to the Kolmogorov-Smirnov (KS) test, the Serine Protease, Tropomyosin, and Mutations external test sets have the most similar length distributions between their allergen and non-allergen sequences (KS statistic $< 0.25$, p-value $> 0.1$, Supplementary Fig. 12b). Consistent with this, on these three test sets, the Length Control strategy consistently outperformed Hard Balance (Fig. 5b). Conversely, on the other external test sets where length distributions were less similar, Hard Balance generally performed better than Length Control, with the exception of Cysteine Protease. These results highlight Length Control as a valuable strategy for building robust models, as it forces them to learn true biological patterns—an advantage that becomes evident on inference tasks lacking such shortcuts. In practice, since test labels are unavailable, applying the Length Control strategy is most likely to be beneficial when the test set is known to comprise sequences of a relatively uniform length.

Finally, to investigate how training set size and imbalance affect model performance on external benchmark datasets, we introduced a "No Balance" strategy. In contrast to our main Hard Balance strategy, No Balance utilized all available sequences, resulting in training sets that were vastly larger but also severely imbalanced due to an overwhelming number of negative instances (Supplementary Fig. 14). These strategies yielded dramatically different class ratios: No Balance was heavily skewed towards negative samples; Hard Balance, Length Control, and Minimal were almost perfectly balanced. The balanced training sets from Length Control and Hard Balance yielded the best-performing models, outperforming No Balance and Minimal (Fig. 5c and d). These findings underscore that for allergen prediction, achieving class balance is more critical than simply maximizing the number of negative training instances. The lower performance of these two strategies likely stems from two distinct causes: whereas Minimal suffered from relatively fewer training data, No Balance was likely overwhelmed by the majority negative class, hindering its ability to learn crucial minority class features.

**Discussion**

In this study, we have presented Applm, a state-of-the-art allergen prediction model that leverages contextualized embeddings from cutting-edge pLMs, including the unprecedented 100B-parameter xTrimoPGLM model. Our work directly confronts two critical shortcomings in the field: an overemphasis on internal CV rather than evaluation in diverse, real-world scenarios, and a lack of stringent similarity-aware benchmarking. These shortcomings have often led to overly optimistic performance claims. To overcome these challenges, we have established a comprehensive evaluation framework. This framework is built upon a curated suite of diverse external benchmark datasets, each targeting a distinct and demanding generalization challenge, and is underpinned by a robust similarity-aware pipeline that ensures fair and reliable assessment. Within this rigorous framework, Applm's superior performance and robustness have been validated through comprehensive settings: 1) extensive testing on both our similarity-aware internal CV and the six external real-world datasets; 2) direct comparison against seven leading methods and conventional protein encodings; and 3) systematic analysis of key factors influencing performance via four distinct dataset construction strategies.

Our findings present a more realistic assessment of allergen prediction capabilities, challenging the exceptionally high performance (e.g., AUROC > 0.9) frequently reported in studies relying on less stringent validation. Such optimistic metrics typically arise from internal CV settings that fail to adequately control for sequence similarity, thereby inflating results. Our own similarity-aware internal CV evaluation underscores this pitfall, revealing that average AUROCs varied widely from 0.722 to 0.951 (Fig. 4c), directly reflecting the influence of sequence similarity. This performance variability became even more pronounced in curated real-world scenarios. On our external benchmark, designed to test challenging tasks such as identifying novel allergens or distinguishing between close homologs and mutants, the model performance was highly context-dependent, with AUROCs spanning a wide range from 0.697 to 0.970 (Supplementary Fig. 5). This complex picture, where models can excel on some external tasks yet show significant vulnerability on others, is the hallmark of a truly realistic evaluation. It demonstrates that continuous innovation, for instance through advanced transfer learning techniques, will be crucial for building models that are not only powerful but also reliably robust across diverse and practical applications.

Beyond allergen prediction, we position our study as a generalizable framework for a wide range of protein classification tasks. The issue of inflated performance metrics due to inadequately controlled sequence similarity is a common pitfall in the field, and our similarity-aware pipeline offers a practical and readily applicable solution. We note that concurrent studies have also begun to address this challenge by introducing similarity-aware data partitioning for training and test sets[46-48,59]. Our work, however, makes a critical extension by controlling for inter-class similarity, a factor often overlooked. Furthermore, we have addressed other confounding factors, such as skewed sequence length distributions, which can impede effective feature learning. We have demonstrated that targeted balancing strategies can yield performance gains in these scenarios. Finally, our allergen prediction results add to the growing evidence of the transformative potential of pLM embeddings for protein informatics. Ultimately, we envision this study as a methodological framework that can inspire future efforts to more rigorously bridge the gap between computational models and their real-world applications.

An emergent debate concerns whether the performance gains from pLMs are partially attributable to a subtle form of "information leakage." It is well-accepted that pLMs avoid traditional data leakage, as their self-supervised training does not access downstream labels. However, this newer concern has emerged regarding whether performance becomes inflated when test proteins are present in the pLM's pre-training data. The evidence on this issue is conflicting. For instance, one study reported a 11.1% performance inflation for protein thermostability prediction[60], whereas another concluded that no performance inflation exists for protein keyword classification[61]. To contribute to this debate, we specifically investigated whether model performance was inflated by pre-training data exposure in the context of allergen prediction. Our By Date external test set serves as an appropriate benchmark, the protein sequences of which were seen during the pre-training of ESM-2, xTrimoPGLM-10B, and xTrimoPGLM-100B, but not ProtT5 (Supplementary text). However, ProtT5 performed comparably to or even better than these three pLMs (Supplementary Fig. 2 and Supplementary Fig. 3). This observation indicates that Applm's predictive performance is robust to pre-training data exposure. Our finding underscores that the impact of pre-training data overlap is likely task-dependent and warrants careful, case-by-case evaluation.

Our study has several limitations. First, while fine-tuning is a promising technique, it did not yield performance gains in our allergen prediction. Future methodological advancements are needed to determine if fine-tuning can be adapted for this specific context or to better understand its inherent constraints. Second, our analysis was confined to sequence-based pLM embeddings. A valuable next step would be to incorporate protein structure-based models, such as those from the ESMFold[33] and AlphaFold[62,63] series, to provide a more holistic understanding. Third, we face a challenge in offering definitive guidance for model selection in real-world scenarios where new datasets are unlabeled. While strategies like Length Control are explored, the optimal model choice is contingent on the target data's characteristics. We therefore propose a heuristic: if a new dataset exhibits high internal similarity (e.g., in sequence length or identity), models trained with our Length Control strategy or on sequences of high inter-class similarity are more likely to yield superior performance. This underscores the need for thorough understanding and careful inspection of the data prior to model deployment. Finally, while this study focused on predictive capabilities, a crucial future direction is to identify feature importance, such as key amino acid positions that drive allergenicity. Such insights would be invaluable for biomedical applications, including the rational design of hypoallergenic proteins that preserve function while minimizing IgE binding.

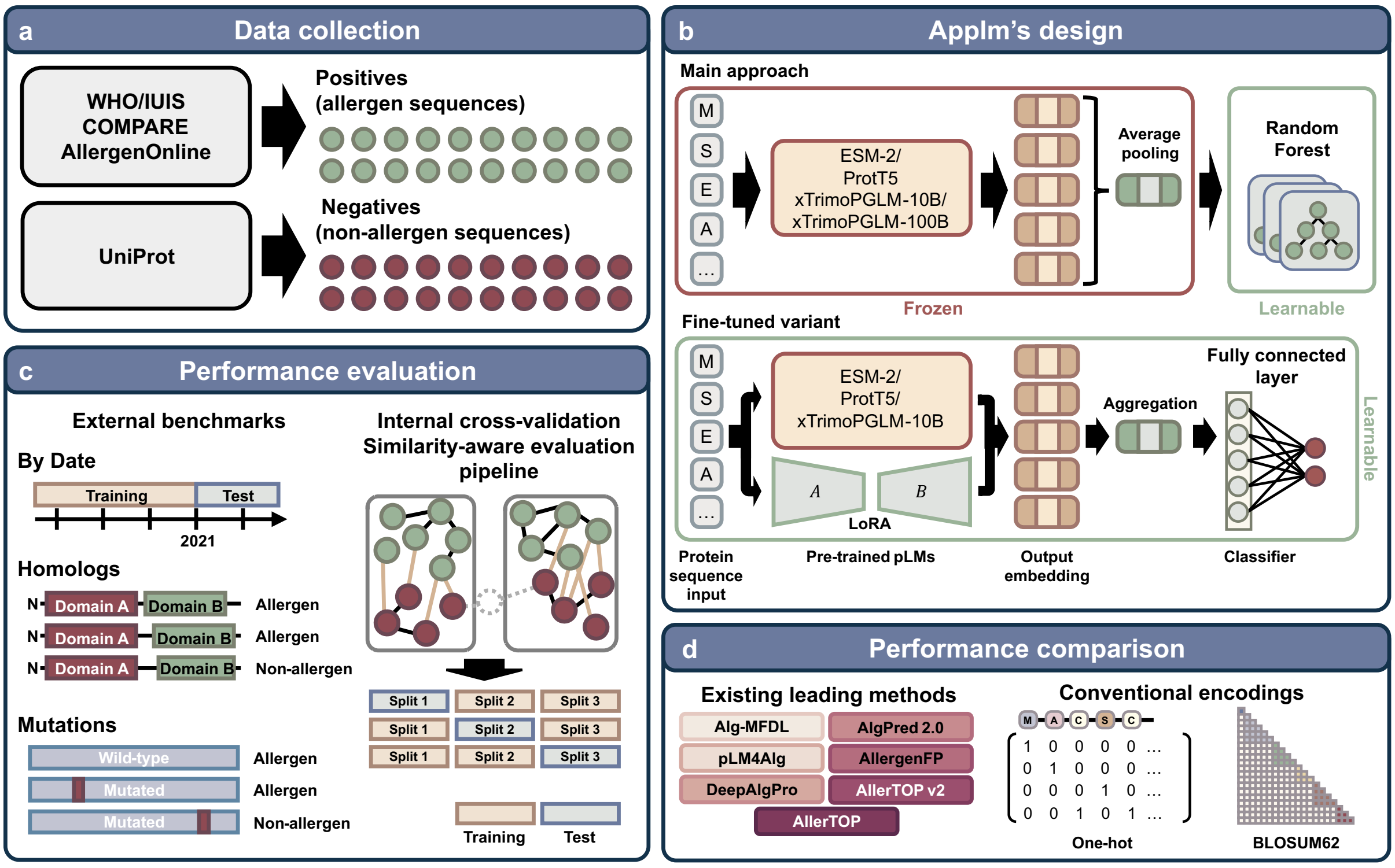

**Fig. 1 Overview of the Applm framework and evaluation methodology. a** Data collection. Allergen sequences were sourced from WHO/IUIS, COMPARE, and AllergenOnline, with non-allergen sequences from UniProt. **b** Applm's design. The main approach uses frozen pLMs to generate protein embeddings for a Random Forest classifier. A fine-tuned variant adapts pLMs using Low-Rank Adaptation (LoRA) and classifies with a fully connected layer that is fed by a vector from an aggregation step, which involves using the classification token from ESM-2, the end-of-sequence token from ProtT5, or average pooling for xTrimoPGLM-10B. **c** Performance evaluation. Internal CV employs our developed similarity-aware pipeline to create training and test splits. External validation is conducted on a benchmark suite composed of a temporal split (By Date), homologs, and mutations. **d** Performance comparison. Applm is benchmarked against seven leading methods and conventional encodings, namely OHE and BL62 encodings.

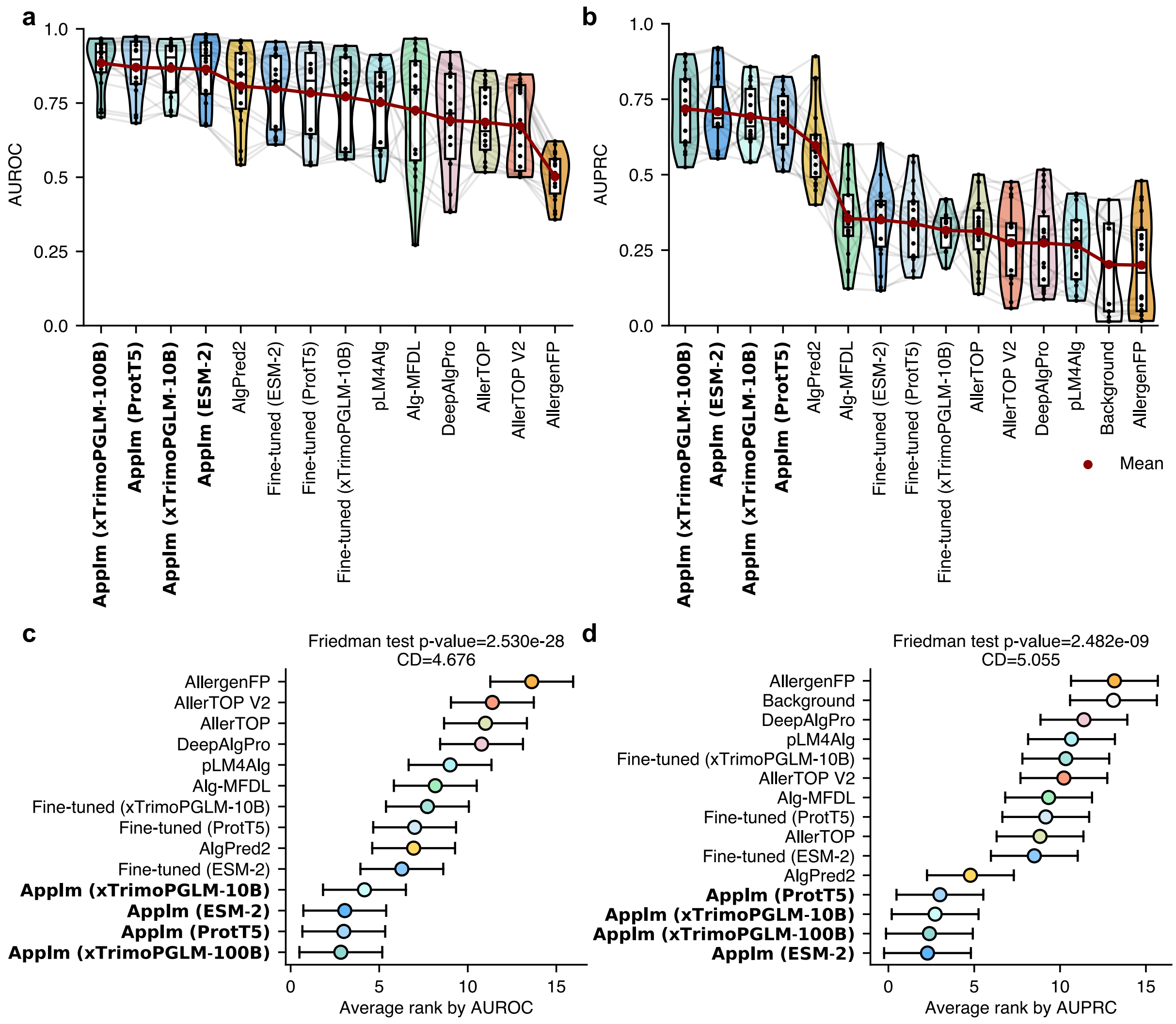

**Fig. 2 Applm outperforms competing methods on the external benchmark. a, b** Distribution of AUROC and AUPRC scores across the external test sets. Models are sorted by their average performance, showing that Applm leveraging four pLMs (bold) consistently achieves the highest scores. **c, d** Average model ranks by AUROC and AUPRC, compared using the Friedman test and the Nemenyi post-hoc test. The Friedman test confirms significant overall performance differences (p-values shown). In the Nemenyi test plots, models whose horizontal lines do not overlap have a statistically significant difference in rank (critical difference, CD, is provided). Applm with different pLMs consistently ranks as the top performer, significantly outperforming nearly all other methods.

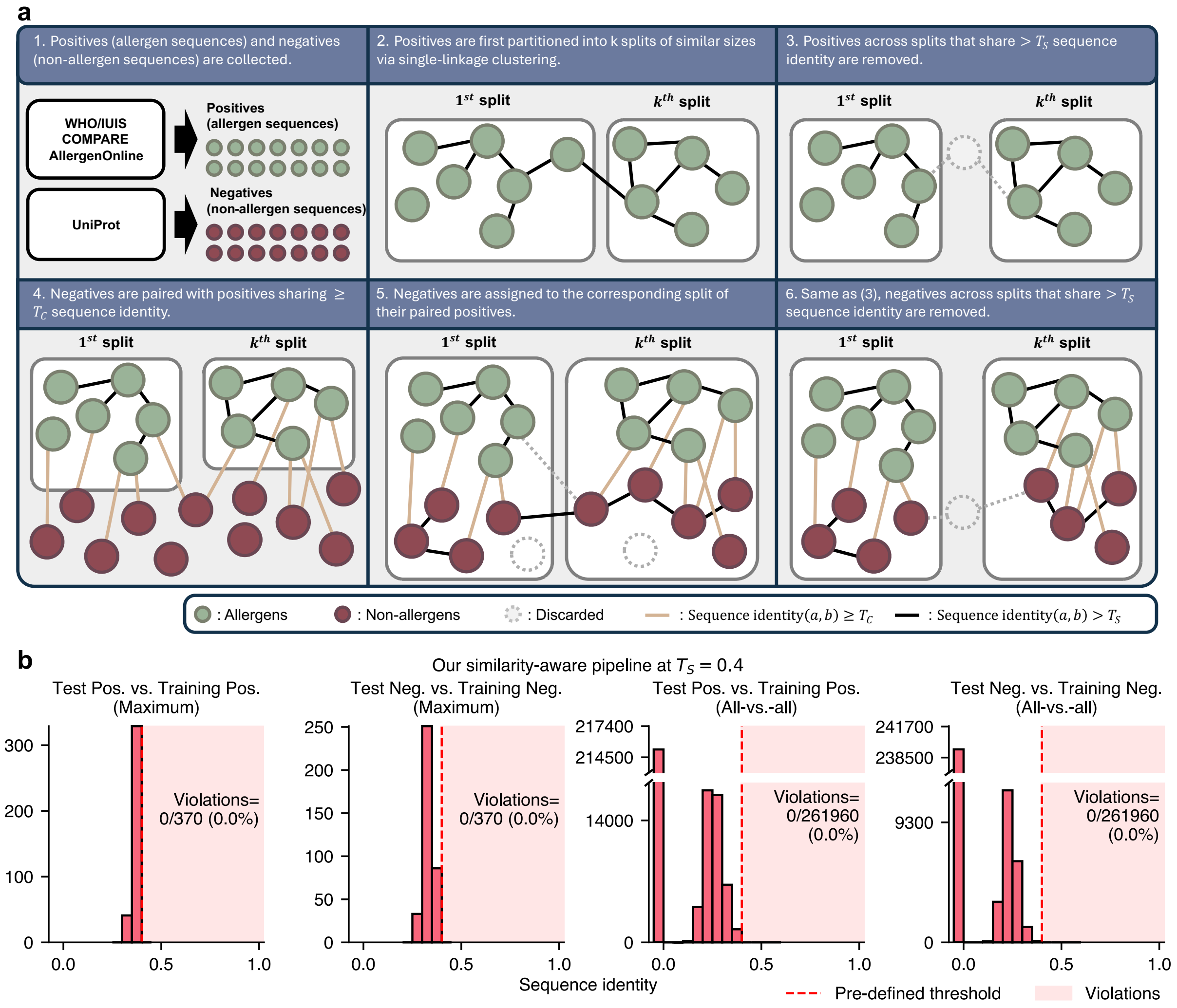

**Fig. 3 Our developed similarity-aware pipeline for robust internal CV. a** A schematic illustrating the pipeline's logic. The example shows a partition into two splits for simplicity; our internal CV setting uses $k = 3$ for 3-fold CV. For visual clarity, the re-addition step for sequences that do not violate the similarity threshold is not shown (Methods). **b** Validation of our pipeline's performance using an inter-split similarity threshold ($T_S$) of 0.4. The histograms show the distribution of maximum and all-vs.-all sequence identity between test and training sets. The results confirm the pipeline's effectiveness, with zero instances exceeding the pre-defined threshold (i.e., zero violations).

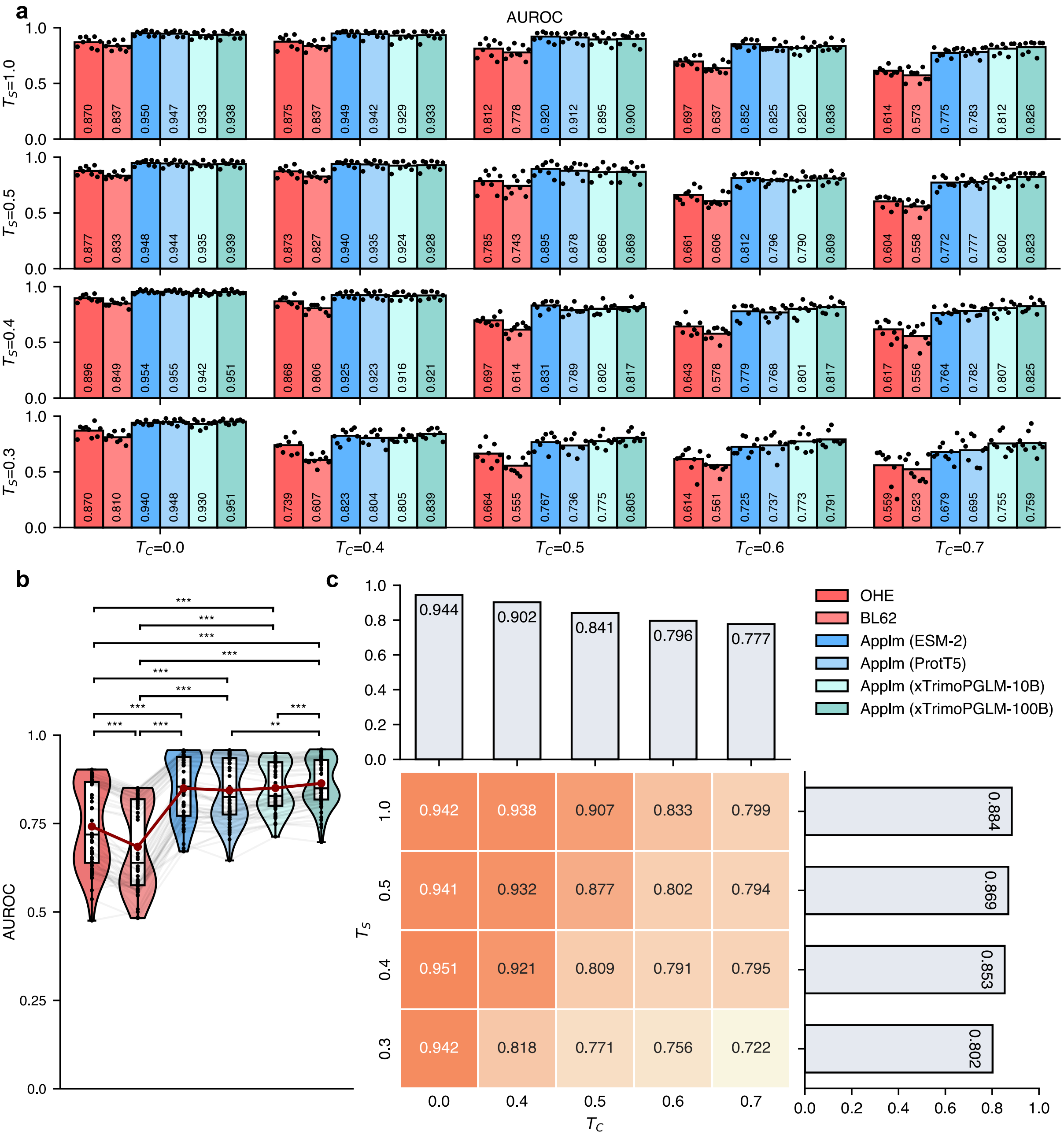

**Fig. 4 Applm outperforms models using conventional encodings on similarity-aware internal CV. a** Bar plots showing the detailed AUROC performance of Applm and models using conventional encodings across a grid of inter-split ($T_S$) and inter-class ($T_C$) similarity thresholds. Each dot represents an individual CV fold. **b** Violin plots comparing the overall AUROC distributions of Applm against models using conventional encodings, aggregated from all conditions in (**a**). **c** Heatmap of AUROC scores, averaged across Applm models leveraging different pLMs, illustrating the combined effect of $T_S$ and $T_C$. The bar plot at the top shows the performance for each $T_C$ level (averaged across all $T_S$ levels), while the bar plot on the right shows the performance for each $T_S$ level (averaged across all $T_C$ levels).

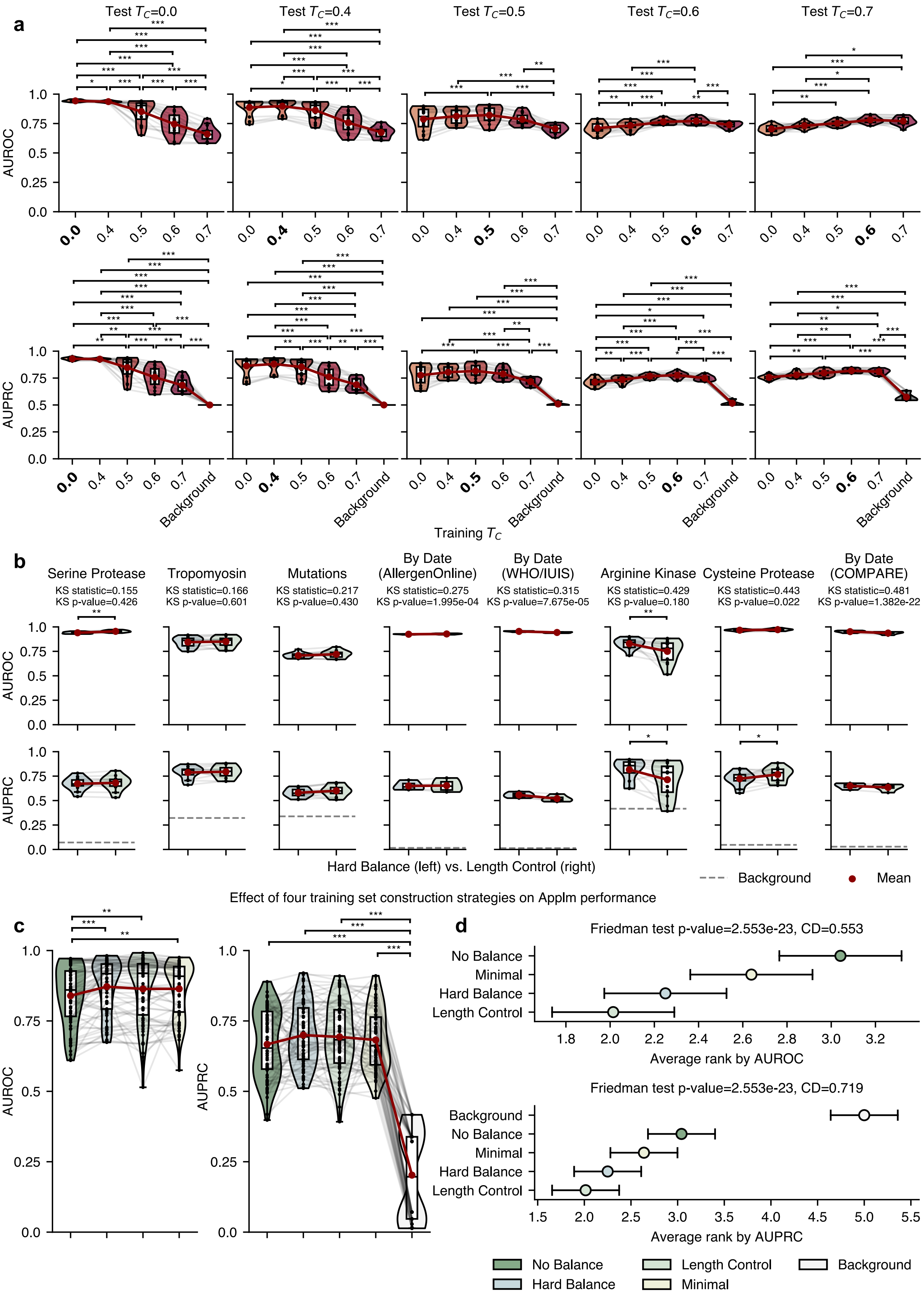

**Fig. 5 Intrinsic difficulty, sequence length distribution, training set size, and training set imbalance impact model performance. a** Performance grid showing AUROC and AUPRC where Applm was trained and tested on datasets with varying inter-class similarity thresholds

($T_C$). For each test $T_C$ (columns), the training $T_C$ (x-axis) that yields the best average performance is highlighted in bold. **b** Performance comparison on each external test set between Applm models trained using the Hard Balance and Length Control strategies. KS test results are shown for each external test set, evaluating the difference in sequence length distribution between positives and negatives. External test sets are ordered by ascending KS statistic. **c** Violin plots comparing the performance distributions of four training set construction strategies, aggregated across all external test sets in the benchmark. **d** Statistical comparison by average rank of the four training set construction strategies shown in (**c**), analyzed with the Friedman test and the Nemenyi post-hoc test.

## Methods

### Applm's pLM embeddings

We utilized four pLM embeddings for Applm: ESM-2, ProtT5, xTrimoPGLM-100B, and its lightweight version xTrimoPGLM-10B.

For ESM-2 and ProtT5, we utilized the 650-million-parameter ESM-2 model (ESM-2_t33_650M_UR50D) and the 3-billion-parameter ProtT5 model (prot_t5_xl_half_uniref50-enc), respectively. Both models were loaded and final-layer embeddings were extracted following their original publications. ESM-2 encoded protein sequences of length L into matrices of size L×1280, while ProtT5 produced matrices of size L×1024.

For xTrimoPGLM-100B and xTrimoPGLM-10B, following a procedure similar to that for ESM-2 and ProtT5, the pre-trained models were loaded and the final-layer hidden embeddings were collected. xTrimoPGLM-100B encoded protein sequences of length L into matrices of size L×10240, while xTrimoPGLM-10B produced matrices of size L×4352.

For each of the four pLMs, we applied average pooling along the sequence length dimension (L) to its output embeddings. This process yields a single, fixed-size vector for each protein, with the vector's dimension being characteristic of the specific pLM used.

### Applm's main approach: frozen pLM embeddings with an RF classifier

We fed the frozen pLM embeddings to an RF classifier for allergen prediction. The RF model was built using the scikit-learn library in Python, with n_estimators set to 1,000 and other parameters left at their default values.

### Applm's exploratory fine-tuned variant: Parameter-Efficient Fine-Tuning (PEFT) with Low-Rank Adaptation (LoRA)

We fine-tuned ESM-2 using LoRA with the Hugging Face transformers and PEFT packages. After an initial hyperparameter search, we decided to use a rank of 4, an alpha of 32, a dropout on LoRA layers of 0.1, a batch size of 1, and 16 gradient accumulation steps. Learning rate started from 5e-7 and linearly increased to 1e-5 in the first 5% of the steps and was subsequently reduced linearly to 9e-6 for the remainder of the training steps. The model was allowed to train for up to 40000 steps and model performance on the validation set was evaluated every 1000 steps.

For the LoRA fine-tuning of ProtT5, we also utilized the Hugging Face transformers and PEFT packages. After an initial hyperparameter search, we settled on a rank of 8, an alpha of 32, a dropout on LoRA layers of 0.1, a batch size of 1, and 16 gradient accumulation steps. Learning rate started from 5e-7 and linearly increased to 5e-5 in the first 5% of the steps and was subsequently reduced linearly to 5e-6 for the remainder of the training steps. The model was allowed to train for up to 95000 steps and model performance on the validation set was evaluated every 1000 steps.

The xTrimoPGLM-10B model was also fine-tuned with LoRA using the same Hugging Face packages. We used a rank of 8, an alpha of 32, a dropout on LoRA layers of 0.1, a batch size of 1, and 16 gradient accumulation steps. Learning rate started from 5e-7 and linearly increased to 1e-5 in the first 5% of the steps and was subsequently reduced linearly to 9e-6 for the remainder of the training steps. The model was allowed to train for up to 15000 steps and model performance on the validation set was evaluated every 1000 steps.

## Data collection and preprocessing for training sets

We gathered allergen protein sequences from three leading and comprehensive databases: WHO/IUIS[11], COMPARE[12], and AllergenOnline[13]. All three databases are peer-reviewed and require experimental evidence before including candidate allergens, but their inclusion criteria differ. For instance, the WHO/IUIS database mandates evidence of specific IgE binding from at least five patient sera, while the COMPARE database only requires IgE binding evidence documented in peer-reviewed studies, with no specified sample size. Due to these differences, we treated each database separately as independent positive sets rather than combining them into a single dataset.

To maintain data consistency and quality, we applied the following quality control criteria to each dataset. Sequences were excluded if they were identical to or substrings of any other sequence. Following previous studies, we also removed sequences shorter than 50 amino acids (AA) or longer than 1000 AA, as well as those containing non-standard amino acids outside of the 20 canonical amino acids. This length range was selected because extremely short sequences may represent truncated recombinant proteins or lack sufficient biological information, while excessively long sequences could distort feature extraction and reduce computational efficiency. After processing, the WHO/IUIS, COMPARE, and AllergenOnline datasets consisted of 1361, 2111, and 2097 allergen sequences, respectively.

For non-allergen protein sequences, we sourced data from UniProt[64], selecting only reviewed eukaryotic proteins that did not contain the allergen tag (KW-0020). This choice was motivated by our observation that almost all allergens from WHO/IUIS originated from eukaryotic sources (1428 allergens from eukaryotes and only 1 allergen from bacteria; Supplementary Fig. 15), suggesting that non-eukaryotic sequences could be trivially classified as non-allergens. We applied the same quality control criteria used for the positive sets to filter the sequences in the negative set. As an additional component of the quality control criteria, we excluded negative sequences that were identical to, substrings of, or contained any positive sequence as a substring. Finally, we retained 164674 non-allergen sequences for the negative set.

## Curating an external benchmark suite of six real-world datasets

For all external test sets, we applied the same quality control criteria for positive and negative sets described in “Data collection and preprocessing for training sets.”

## Four homolog external test sets

We curated four external protein homolog test sets for protein families known to include allergens: Arginine Kinases[65,66], Cysteine Proteases[67,68], Serine Proteases[69,70], and Tropomyosin[71]. For each family, we obtained allergen sequences from the WHO/IUIS database, which provides experimental evidence of allergenicity. We retrieved non-allergen sequences from UniProt by selecting reviewed proteins without allergenicity annotations. To improve the reliability of the non-allergen set, we excluded sequences lacking the specific functional annotation ("EC 2.7.3.3" for arginine kinase, "EC 3.4.22" for cysteine protease, and "EC 3.4.21" for serine protease). No annotation filtering was applied for tropomyosin, as it is not an enzyme. It is important to note that the absence of allergenicity evidence does not confirm a protein as a non-allergen, it only indicates that no allergenicity studies have been reported. To address this uncertainty, we performed a final manual curation based on biological knowledge described below, ensuring the robustness and biological relevance of each homolog test set.

**Arginine Kinase**

Studies have indicated that arginine kinases from crustaceans in the Decapoda order are frequently allergenic[65,66]. Cross-reactivity between arginine kinases from shrimp and crab species, both within Decapoda, has also been reported[72]. To reduce the high potential for cross-reactivity, we excluded all non-allergen sequences from Decapoda, even in the absence of direct experimental evidence of allergenicity. After filtering, the arginine kinases external test set contained 10 allergens and 14 non-allergens.

**Cysteine Protease**

The cysteine proteases external test set contained 11 allergens and 223 non-allergens.

**Serine Protease**

Studies have shown that repeated exposure to snake venom can trigger IgE-mediated, allergy-like immune responses[73,74]. Serine proteases have been identified as components of snake venom allergy[75,76]. To address this, we excluded all serine proteases from the Viperidae family (snakes) from the non-allergen set. After filtering, the serine proteases external test set contained 32 allergens and 417 non-allergens.

**Tropomyosin**

Among tropomyosin non-allergens, two sequences from *Blattella germanica*[77] (German cockroach) and *Periplaneta fuliginosa* (Smokybrown cockroach) were excluded, as tropomyosin from cockroach species such as *Blattella germanica* and *Periplaneta americana*[78] (American cockroach) have been reported as allergens. After filtering, the Tropomyosin external test set contained 28 allergens and 59 non-allergens.

**The "By Date" external test set**

To simulate real-world scenarios where newly discovered sequences need accurate predictions, we constructed external test sets for each of the three datasets based on sequence entry creation dates. Specifically, we isolated all sequence entries created after 2020 in each database as external test sets, while retaining entries created on or before 2020 as the training set. For negative sequences, we performed the same approach: entries created in UniProt after 2020 were assigned to the negative test set, while those created on or before 2020 were included in the training set. The By Date external test sets for WHO/IUIS, COMPARE, and AllergenOnline contained 50, 107, and 60 allergens, respectively, and 3655 non-allergens each.

### The "Mutations" external test set

We curated an additional external test set from published allergen mutational scanning experiments[79-91] (Supplementary Table 2). These studies explored the allergenicity of allergens after inducing mutations at one or more positions. We labeled the mutated allergens as either positive or negative based on the allergenicity reported in their respective studies, while the original wild-type allergens were labeled as positives. This dataset was designed to present a significant challenge due to the high sequence similarity between positive and negative samples. The final Mutations test set included 22 allergens and 43 non-allergens.

### Comparing Applm's performance with seven published methods on the real-world external benchmark

We first constructed training sets for each of the external benchmark datasets described above. Each positive dataset (WHO/IUIS, COMPARE, AllergenOnline) and the UniProt negative dataset served as the starting point for creating the training set. For the By Date external test set, sequence entries created on or before December 31, 2020 (i.e., on or before 2020), were used for the training set. Sequences from the training set that were identical to, a substring of, or a superstring of any sequence in each corresponding external test set were removed.

We then adopted our main training strategy, "Hard Balance," aiming to match the number of positives and negatives in the training set as closely as possible by down-sampling the majority negative class, enabling evaluation in a label-balanced scenario. To create each validation split, we randomly sampled 10% of sequences (or up to 500 sequences, whichever was smaller) from each training set while maintaining the positive-to-negative ratio. The remaining sequences were used as the final training split.

The final training split was used exclusively for model training. The validation split was reserved for hyperparameter tuning, model selection, and early stopping for some models (details below). Specifically, for Applm's main approach, the validation split was not used. For Applm's exploratory fine-tuning of ESM-2, ProtT5, and xTrimoPGLM-10B, the validation split was used to compute the validation loss and determine early stopping within the allowed fine-tuning steps range described above. For these fine-tuned models, the checkpoint with the lowest validation loss was selected for final evaluation.

We reproduced and tested seven published models on our external test sets: AllergenFP[19], AllerTOP[16], AllerTOP v2[17], AlgPred2[21], DeepAlgPro[26], pLM4Alg[25], and Alg-MFDL[32]. These models were selected based on the availability of code or clear instructions for direct reproduction provided in the respective publications. Each model was re-trained using the same training set as Applm to ensure fair and consistent comparisons. A brief overview of their implementation is provided below.

### AllergenFP

AllergenFP predicts allergenicity by comparing the computed fingerprints of proteins to those of known allergens and non-allergens, assigning each protein the label of its closest match. Fingerprints were generated using E-descriptors and auto-cross covariance (ACC) transformations, with similarity between sequences calculated via Tanimoto coefficients. Consistent with the original study, we classified each test protein based on its nearest training instance.

**AllerTOP**

AllerTOP predicts allergenicity using protein physicochemical properties. Protein sequences were encoded using Z-descriptors and ACC transformations, and classification was performed using a k-NN algorithm. All hyperparameters were set to the values specified in the original publication.

**AllerTOP v2**

AllerTOP v2 is an updated version of AllerTOP, where Z-descriptors were replaced with E-descriptors. All hyperparameters were set to the values specified in the original publication.

**AlgPred2**

AlgPred2 combines predictions from three components: 1) an RF model trained on amino acid composition, 2) a motif search using MERCI on a predefined set of allergen epitopes, and 3) a BLAST search against a training set of allergens and non-allergens. We strictly followed the original methodology and trained RF models exclusively on the training split. We also created BLAST databases for each training split using makeblastdb from the BLAST package (version 2.16.0+). The predefined motif set was used without modification, and motif search was performed using the provided MERCI Perl script. Predictions were generated using the provided inference script by AlgPred2.

**DeepAlgPro**

DeepAlgPro employs a deep learning architecture that combines a convolutional layer and a self-attention layer. We obtained the model architecture and training scripts from the original repository. Following the original implementation, we trained the model for up to 100 epochs, and the checkpoint with the lowest validation loss was selected for final evaluation on external test sets. Default hyperparameters were used without modification.

**pLM4Alg**

pLM4Alg is a deep learning model that utilizes convolutional and pooling layers to process ESM-2 encoded protein sequences as input. The optimal hyperparameter values reported by the original publication were applied, including a convolutional layer with 32 channels, a dense layer with 4096 neurons, a kernel size of 9, and a stride size of 2. ESM-2 with 150 million parameters (ESM-2_t30_150M_UR50D) was used for encoding as is used in their web server. Following their implementation, we trained the model for up to 100 epochs, and the checkpoint with the lowest validation loss was selected for final evaluation on external test sets. During reproduction, we observed substantial training instability using their provided learning rate (0.1). To address this, we systematically tested smaller learning rates (0.01, 0.001, 0.0001), which demonstrated more stable and improved performance on the validation split. Therefore, we ultimately trained four models on each training setting with these learning rates and the model trained on the learning rate with the lowest validation loss was selected for final evaluation on external test sets.

### Alg-MFDL

Alg-MFDL is a deep learning model that encodes protein sequences using ESM-2, ProtT5, Protein Position-Specific Scoring Matrix (PSSM), and Dipeptide Deviation from Expected Mean (DDE). The model uses the 650-million-parameter version of ESM-2 and the 3-billion-parameter version of ProtT5 to encode protein sequences. Various model architectures were explored in the original study, and we adopted the one with the best reported performance, which combines convolutional and pooling layers. Following the original implementation, we computed PSSM using psiblast from the BLAST package[92] (version 2.16.0+) and DDE using the provided script, while ESM-2 and ProtT5 embeddings were obtained as previously described. For PSSM encoding, the original study used UniRef50 as the database to estimate amino acid frequencies. However, generating PSSM encodings for our full training and test sets would have required approximately 35 days due to the computational demands of multiple sequence alignment (MSA). To address this challenge, we created a smaller subset of the UniRef50 database by randomly selecting a subset containing two orders of magnitude fewer sequences to use for calculating PSSMs. This reduced the PSSM encoding time to approximately 16 hours. After encoding, we concatenated the representations from ESM-2, ProtT5, PSSM, and DDE, following the original implementation. Models were trained for up to 100 epochs, and the model that achieved the highest validation accuracy was selected for final evaluation on our external test sets.

### FFNN as the classifier

Additionally, we evaluated an alternative classifier by replacing Applm's RF with an FFNN and comparing their performance. The FFNN, implemented in PyTorch, was constructed with a single hidden layer of 1,024 neurons. To mitigate overfitting, the hidden layer included L2 regularization (weight decay of 1e-6), batch normalization, and dropout (rate = 0.5). We trained the model for up to 100 epochs using a learning rate of 1e-3, with early stopping to select the checkpoint that achieved the lowest validation loss.

### Calculation of AUROC and AUPRC

We used AUROC and AUPRC to evaluate model performance. AUROCs and AUPRCs were calculated in R using PRROC, a reliable tool proven to compute these metrics correctly[93]. Background AUPRC was calculated as the ratio of the number of actual positive instances to the total number of instances.

### Developing a similarity-aware pipeline for clean partitioning of protein datasets

In this study, we developed a novel framework to partition a two-class dataset into $k$ distinct splits while maintaining critical similarity constraints. Specifically, we controlled inter-split similarity by ensuring that no sequences across splits shared an identity greater than a predefined threshold ($T_S$). At the same time, we controlled inter-class similarity by requiring that within each split, negative sequences shared an identity greater than or equal to a predefined threshold ($T_C$) with at least one positive sequence.

**Smith-Waterman local alignment identity**

Before clustering, we quantified the similarity between sequences using the Smith-Waterman local alignment identity[94]. This was calculated with the ssearch36 function from the fasta36 package[95], using the BL62 scoring matrix and an E-value threshold of 1e10 to ensure all pairwise alignments could be computed and returned. Default settings were used for all other parameters. We applied a coverage control: if the alignment length was less than 25% of the shorter sequence's length, the alignment identity was set to 0. This step helped reduce the impact of extremely short regions that could produce spurious alignments with high sequence identity. Pairwise sequence identities for all sequences were calculated and stored for clustering. Unless otherwise stated, all sequence identities in this study were computed following this procedure.

**Partitioning the positive set**

Our splitting strategy began by dividing the positive sequences into $k$ splits, followed by partitioning the negative sequences. Each positive dataset $D \in \{D_{WHO/IUIS}, D_{COMPARE}, D_{AllergenOnline}\}$ was processed independently. For each dataset, the positives were divided into three roughly equal splits. Specifically, given the desired number of splits $k$ ($k = 3$ for 3-fold CV in this study), the maximum size of each split was limited to $S = \lceil\frac{|D|}{k}\rceil$. Positive sequences were iteratively assigned to splits using a single-linkage clustering algorithm. A list of positive sequence pairs, ordered by sequence identity, was generated. Starting with the pair of the highest identity, sequences were grouped into clusters. If a grouping step would result in a cluster exceeding the size limit $S$, that step was skipped. Clustering concluded once all pairs in the list had been processed, resulting in three splits with highly similar sizes. In cases where the size limit causes the algorithm to produce more than three clusters, the smallest clusters were iteratively merged until only $k$ splits remained, allowing size violations as necessary while keeping target split sizes as close as possible.

After clustering, the $k$ splits were checked for inter-split sequence pairs that shared $> T_S$ sequence identity ($T_S \in \{0.3, 0.4, 0.5, 1.0\}$). For every sequence pair across splits that shared $> T_S$, the total number of inter-split violations was calculated as the sum of violations for both sequences. For example, the total number of violations for sequence $S_{A1}$ from split $A$ against any sequence in splits $B$ and $C$ was recorded as $V_{S_{A1}}$, and $V_{S_{B1}}$ was similarly calculated for $S_{B1}$ from split $B$. The total violations for the pair $P_{S_{A1},S_{B1}}$ were calculated as $V_{S_{A1}} + V_{S_{B1}}$. These values were efficiently obtained using a precomputed distance matrix.

To resolve inter-split violations, we first prioritized pairs with the highest total violations. For every pair of inter-split violations, the sequence belonging to the larger split was removed; if both splits had the same size, a sequence was removed at random. Each removed sequence and its number of violations was recorded for a later step. This strategy prioritized removing sequences with high connectivity, minimizing the overall number of removals required, while maintaining size balance between splits. We also explored an alternative approach where sequences (rather than pairs) were removed starting from sequences with the most violations. In practice, this method often disproportionately removed sequences from one or two splits, causing split sizes to become severely imbalanced. Consequently, we settled on our current approach of ranking sequence pairs. We iterated through all inter-split violation pairs, while recording removed sequences and their violation counts.

After all violations were removed, we attempted to add back sequences starting with those possessing the least number of recorded violations so that more sequences could be retained without triggering any violations. Starting from the least number of violations, removed sequences were iteratively added back to their original splits only if the addition did not introduce any inter-split violations. In practice, we found that this re-addition step retained roughly 20% of sequences that would otherwise be removed. This final re-addition step optimized split sizes while maintaining a clean separation between splits.

**Partitioning the negative set**

We used a pairing approach to construct the negative splits. Negative sequences were paired with positives if they shared a sequence identity $\geq T_C$ and were subsequently assigned to the split corresponding to their paired positive. Starting with the highest threshold ($T_C = 0.7$), negative sequences sharing $\geq 0.7$ identity with any positive sequence in any split were identified. Negatives were first assigned if they were the only pair to a single positive sequence. For the remaining negatives, the number of possible splits each negative could be paired with was recorded. Assignment then proceeded by starting with negatives that could be paired with only one split. Then, if a negative sequence could be assigned to multiple splits, it was assigned to the split with the fewest negatives at the moment of assignment. After all assignments, negative sequences were checked for inter-split pairs sharing $> T_S$, and any such sequences were removed as described previously.

For subsequent lower $T_C$ thresholds, we partitioned the negatives using splits from a previous higher $T_C$ as a foundation rather than starting from scratch. Additional negative sequences meeting the current $T_C$ threshold with any positives were first identified. The eligible negative sequences were then assigned to positive splits, starting with negative sequences that could be assigned to the fewest splits. Before each assignment step, the negative sequence was checked to see if it shared $> T_S$ identity with negative sequences in other splits. For example, when creating negative splits for $T_S = 0.4$ and $T_C = 0.6$, the negative splits from $T_S = 0.4$ and $T_C = 0.7$ served as the starting point. Negatives that could be paired with positives at $T_C = 0.6$ but not at $T_C = 0.7$ were added to the appropriate split as described above, with violations checked at each assignment step. In practice, we found that the number of new negatives that could be paired between successive $T_C$ thresholds was small enough that checking for violations at each assignment step was more efficient than performing them after all assignments. While negative sequences can be partitioned independently at any $T_C$ threshold, we observed that using splits from previous $T_C$ as a starting point would produce more consistent partitions and results. A total of $3 \times 4 \times 5 = 60$ 3-fold CV sets were created using the three datasets, 4 $T_S$ levels, and 5 $T_C$ levels ($T_C \in \{0.0, 0.4, 0.5, 0.6, 0.7\}$).

**Constructing four strategies for model training**

The original splits generated by our pipeline described above constituted the No Balance strategy. Hard Balance and Length Control strategies were then created from each split. In the Hard Balance strategy, the negative set in each split was randomly subsampled to match the size of the respective positive set. If the negative set contained fewer sequences than the positive set, no subsampling was performed.

In the Length Control strategy, the negative set was subsampled to match not only the size but also the sequence length distribution of the positive set. This ensured that the two sets not only had equal sizes but also comparable sequence length distributions. As with the Hard Balance strategy, no subsampling was performed if the negative set contained fewer sequences. Consequently, Hard Balance and Length Control strategies would produce identical splits when no subsampling was applied.

Finally, the Minimal strategy was implemented to standardize the total size and class distribution across all data splits. This process began by identifying the minimum number of positive and negative examples present across all previously generated splits. For external benchmarking, this was a single global minimum found across all settings, whereas for internal CV, a separate minimum was determined for each dataset individually (WHO/IUIS, COMPARE, and AllergenOnline). With these target minimums established, the Minimal setting was generated for each existing $T_S$ and $T_C$ condition. First, the positive examples were randomly subsampled down to their defined minimum count. Subsequently, the negative set was constructed by sampling exclusively from a pool of candidates that shared a sequence identity of $\geq T_C$ with this new, smaller set of positives, continuing until the defined minimum for negatives was also met.

Throughout our study, we applied Hard Balance as the main strategy for both external benchmarking and internal CV. For internal CV, we also conducted experiments using the Minimal strategy (Supplementary Fig. 8, Supplementary Fig. 9, and Supplementary Fig. 10). For external benchmarking, we compared the Hard Balance strategy with the No Balance, Length Control, and Minimal strategies (Fig. 5b, c, and d).

**Visualizing sequence identity distribution**

To analyze sequence similarity, we characterized the all-vs.-all sequence identity distribution between two sets of sequences. Pairwise sequence identities were calculated using local alignment sequence identity, as described earlier. These pairwise identities, representing the similarity between every sequence in one set and every sequence in the other, were visualized as histograms to display the distribution.

In addition to the all-vs.-all identity distribution, we visualized the maximum sequence identity between the two sets. For each sequence in one set, we calculated the highest sequence identity it shared with any sequence in the other set. These maximum sequence identities were also visualized as histograms. This distribution allows a clearer representation of the closest sequences between two sets.

**Conventional protein sequence encoding**

OHE encodes each amino acid as a one-hot vector of size 1×21, with a value of 1 assigned to the position corresponding to one of the 20 standard amino acids or a single placeholder for unknown residues. Protein sequences of length L were represented as matrices of size L×21.

BL62 maps each amino acid to a vector of size 1×23, corresponding to the respective column in the BLOSUM62 substitution matrix[96]. Protein sequences of length L were encoded as matrices of size L×23.

**Investigating the impact of "difficulty matching" on model performance**

For internal CV, we designed additional experimental settings using separate $T_C$ thresholds for the training and test sets. For all the previously generated internal 3-fold CV splits, each split was used as the test split, while all remaining sequences were combined to form the training split. Within the training split, sequences were filtered to ensure no sequence shared $> T_S \in \{1.0, 0.5, 0.4, 0.3\}$ identity with any sequence of the same class in the test split. Subsequently, negatives in the training split were retained only if they shared $\geq T_C \in \{0.0, 0.4, 0.5, 0.6, 0.7\}$ identity with at least one positive sequence in the same training split. For example, to create a setting where $T_S = 0.5$, training $T_C = 0.4$, and test $T_C = 0.5$, we began with the three splits from the original internal CV setting where $T_S = 0.5$, $T_C = 0.5$ using No Balance. Each of the three splits was assigned as the test split once, and every remaining positive and negative sequence not part of the test split was assigned as training sequences. Training sequences were first filtered out based on $T_S = 0.5$ with the test split. Then, five training splits were created at training $T_C \in \{0.0, 0.4, 0.5, 0.6, 0.7\}$, including training $T_C = 0.4$. The training splits were then subsequently constructed for Hard Balance and Minimal as described above. This experiment was performed on the WHO/IUIS dataset.

For external benchmarking, in addition to our main setting, we also performed experiments at $T_S = 0.5$ and $T_C \in \{0.0, 0.4, 0.6\}$ to increase the diversity for comparing intrinsic task difficulty between training and test sets. Specifically, in the $T_S = 0.5$ setting, sequences from the training set sharing $> 0.5$ sequence identity to any sequence of the same class in each respective external test set were removed. Next, for both the training sets constructed from our main setting and the $T_S = 0.5$ setting, we applied three inter-class similarity restrictions ($T_C \in \{0.0, 0.4, 0.6\}$) to each training set as described earlier. Subsequently, we characterized each training set and external test set by calculating the maximum similarity of each negative sequence to every positive sequence within the respective sets. For each pair of training and external test sets, we used the Mann-Whitney U[97] test to compare the similarity distributions and determine if the inter-class similarity in the training set and the external test set differed significantly. Finally, we grouped the performance of Applm in different settings based on whether the training set had a significantly higher inter-class similarity than the external test set, a significantly lower inter-class similarity, or no significant difference.

**Investigating the impact of sequence length distribution on model performance**

To explore the impact of sequence length on model performance, the sequence length distributions of positive and negative sequences in each dataset and external test set were calculated and compared using the KS test[98]. The KS statistic and corresponding p-value were calculated and recorded.

**Statistical tests**

For all statistical tests, unless otherwise specified, we used the non-parametric two-sided paired Wilcoxon signed-rank test[99] to compare distributions of continuous values. For unpaired performance comparisons, we applied the Mann-Whitney U test. All p-values from multiple comparisons were adjusted using Bonferroni correction. Throughout this study, significance levels are indicated by asterisks as follows: $^{*}p < 0.05$, $^{**}p < 0.01$, and $^{***}p < 0.001$, unless otherwise specified. To compare model performance across multiple settings, we employed the Friedman test[100] and the Nemenyi post-hoc test[101]. The Friedman test is a non-parametric statistical test that compares the performances of multiple models across multiple test sets. Instead of analyzing raw scores, it evaluates whether there are significant differences in the ranks of the models' performances, making it robust to non-normal distributions and outliers. If the Friedman test identified significant differences, the Nemenyi post-hoc test was applied to determine which specific models differed. The Nemenyi test calculates the critical difference (CD), which is the minimum difference in average ranks required for two models to be considered significantly different. Models whose rank differences exceed the CD are deemed to have statistically distinct performances.

## Data and code availability

All data used in this study can be obtained following the procedure described in the Methods section. The code and processed data in this study are available in the Applm repository on GitHub at https://github.com/brianwongsh/Applm.

## Acknowledgments

QC is supported by National Natural Science Foundation of China under Award Number 32100515 and CUHK direct grant for research under Award Numbers 2022.080 and 2025.031.

## Contributions

BSHW and QC conceived the project. KYY, SKWT, and QC supervised the project. BSHW and QC designed the computational experiments and data analyses. BSHW and JMK prepared the data. BSHW implemented the methods, conducted the experiments, and analyzed the results. JMK and SHF independently reproduced parts of the results. All authors interpreted the results. BSHW, KYY, and QC wrote the manuscript. All authors reviewed and approved the final manuscript.

## Corresponding authors

Correspondence to Kevin Y. Yip, Stephen Kwok-Wing Tsui or Qin Cao.

## Ethics declarations

No ethical approval was required for this study. All utilized public datasets were generated by other organizations that obtained ethical approval.

## Competing interests

The authors declare that they have no competing interests.

**Supplementary text**

**Clustering tools such as CD-HIT cannot ensure desired similarity separation**

CD-HIT, BLASTClust[92], and the more recent MMseqs2[102] are widely used clustering tools originally designed to group similar protein sequences. CD-HIT and MMseqs2 ensure that each sequence within a cluster meets the similarity threshold with the cluster's representative sequence, while BLASTClust enforces pairwise similarity between all sequences within a cluster. However, none of these tools explicitly enforce the thresholds on similarities between clusters. Therefore, these tools are not reliable for partitioning datasets in tasks requiring strict separation of training and test sets based on sequence similarity. Recent studies have also corroborated similar misuse of these tools in other protein sequence prediction tasks[35,45-48].

We examined whether training and test sets from previous studies contained sequences exceeding a specified similarity threshold (i.e., violations). We collected the training and test sets from AlgPred2, which reported using CD-HIT with a 0.4 threshold, and generated two additional training and test sets using CD-HIT and MMseqs2 with the same threshold (details below in "Partitioning and checking sequence similarity using CD-HIT" and "Partitioning and checking sequence similarity using MMseqs2"). We quantified sequence similarity by computing all-vs.-all identity and maximum identity between training and test sets (Methods). For each test sequence, all-vs.-all identity measures similarity with all training sequences of the same class, while maximum identity captures the highest similarity with any training sequence of the same class. As expected, in all three scenarios, test sequences showed substantial similarity to training sequences, frequently exceeding the desired threshold (Supplementary Fig. 1). Maximum identity revealed violations more obviously than all-vs.-all identity, indicating potentially inflated performance, as methods could easily rely on most similar training sequences for predictions.

**Partitioning and checking sequence similarity using CD-HIT**

We partitioned the dataset using the CD-HIT (version 4.8.1) package at an inter-split identity threshold of $T_S = 0.4$. First, positive and negative sequences were clustered separately with the clustering threshold set to 0.4 (-c 0.4). Local alignment was enabled (-G 0), and a minimum alignment coverage of 25% was required (-aS 0.25). The resulting clusters were then sorted by decreasing size and iteratively assigned to the smallest of the three splits to ensure the splits were approximately balanced. We then randomly merged two splits to create the training set and used the remaining split as the test set. To assess the effectiveness of this partitioning, we then used the cd-hit-2d utility from the same package to check for similarities between the training and test sets. However, it is critical to note that this verification method itself is not exhaustive. The cd-hit-2d utility relies on internal heuristics (e.g., a word-size filter) and is not designed to report every possible pairwise alignment between two sets. Consequently, the number of violations detected by this method is likely an underestimation of the true number of sequences that exceed the 0.4 similarity threshold between the training and test sets.

For the splits provided by AlgPred2, we directly used cd-hit-2d to check sequence similarity as described above, because splits used in AlgPred2 were also created using CD-HIT.

**Partitioning and checking sequence similarity using MMseqs2**

We also partitioned the dataset using MMseqs2 (version 15.6f452) at an inter-split identity threshold of $T_S = 0.4$. Following a similar procedure to the one used for CD-HIT, positive and negative sequences were clustered separately via the mmseqs easy-cluster command. Key parameters included setting the minimum sequence identity to 0.4 (--min-seq-id 0.4), a coverage threshold of 0.25 (-c 0.25), and coverage mode 1 (--cov-mode 1). The resulting clusters were allocated to three splits and combined to training and test sets as described above. To assess the effectiveness of this partitioning, we used the mmseqs easy-search command to check for similarities between the resulting training and test sets. However, it is important to note that, much like CD-HIT's similarity search tool, this verification method is not exhaustive. The mmseqs easy-search command employs a fast, heuristic-based search that is not designed to guarantee finding every possible pairwise alignment. Although we configured the search to be highly permissive (e.g., --min-seq-id 0, -e 1E10), the algorithm's fundamental heuristics remain active. Consequently, the number of violations reported by this method is also likely an underestimation of the true number of sequences that exceed the 0.4 similarity threshold between the training and test sets.

**Rationale for curating the real-world external benchmark**

Among the few previous studies incorporating external validation, a common practice was the use of "By Date" external test sets. These sets reflect the real-world scenario of classifying newly discovered or characterized proteins.

The second benchmark scenario focuses on the challenge of distinguishing allergens within homologous protein families. This scenario is crucial for evaluating allergen risks in food (e.g., genetically modified foods, novel food sources) and environmental exposures, underscoring the difficulty of identifying allergenic proteins when they share high sequence similarity with non-allergenic family members. For instance, while seafood tropomyosin is a common allergen, its mammalian and bird counterparts are generally not allergenic[71].

The third benchmark scenario involves evaluating mutated variants of known allergens. This scenario depicts the significant challenge of predicting the allergenicity of protein variants differing by only one or a few amino acids from wild-type allergens. Such precise discrimination is essential for designing hypoallergenic proteins that reduce IgE binding while preserving function, for ensuring the safety of genetically engineered proteins, and for detecting risks from processing-induced changes.

**Robustness of pLMs' performance to pre-training data exposure**

The cutoff date for ProtT5's pre-training data is on or before 2020, a detail we confirmed via private communication with the authors as it was not specified in the original publication.

| Year published | No. | Method name | Encoding | Variable length handling | Allergen sources | Non-allergen sources | No. of allergens | No. of non-allergens | Inter-split (training vs. test) similarity control | Inter-class (allergen vs. non-allergen) similarity control | Length control | Internal test sets | External test sets | Model | Reported performance: Acc | Sens | Spec | Prec | F1 | MCC | AUROC | AUPRC (Background) |
|---|---|---|---|---|---|---|---|---|---|---|---|---|---|---|---|---|---|---|---|---|---|---|
| 2025 | 1 | Alg-MFDL | Frozen pre-trained embeddings from ESM2 and ProtT5, AAC-PSSM, and DDE | Aggregated by encoding (average pooling) | SDAP, COMPARE, AllergenOnline, WHO/IUIS | UniProt | 3550 | 3550 | NA | Non-allergens with a mutual sequence similarity of over 40% and a similarity to allergens greater than 40%, as calculated by CD-HIT, were removed | NA | 710 allergens and non-allergens were randomly sampled from the training set to form the test set | None | CNN | 0.973 | 0.951 | | 0.996 | 0.973 | | | |
| 2024 | 2 | SEP-AlgPro | Embeddings from ProtT5, ESM1b, ESM1v, ProtBERT, ProtAlbert, Seq2Vec, PLUSRNN, Bepler | Aggregated by encoding (average pooling) | SDAP, WHO/IUIS, AllergenOnline, COMPARE, UniProt | UniProt | 2820 | 2820 | NA | Non-allergens with a mutual sequence similarity of over 40% and a similarity to allergens greater than 40%, as calculated by CD-HIT, were removed | NA | 1) 710 allergens and non-allergens were randomly sampled from the training set to form the test set<br>2) A test set was created with 550 allergens and 4769 non-allergens, which were sourced by merging training sequences from AlgPred2 and NetAllergen and then removing sequences that were redundant with the training dataset<br>3) A dataset was compiled using 171 allergens and 1810 non-allergens from AlgPred2's external test sets, after excluding sequences that were similar to those in the training set<br>4) A dataset was created using 928 allergens and 4740 non-allergens from the NetAllergen dataset, after excluding sequences that overlapped with the training set | None | DNN, GBT, LGBT, ABT, XGBT, SVM, CB, prediction results integrated with a final DNN. | 1) 0.957<br>2) 0.937<br>3) 0.974<br>4) 0.941 | 1) 0.938<br>2) 0.958<br>3) 0.906<br>4) 0.901 | 1) 0.976<br>2) 0.935<br>3) 0.980<br>4) 0.949 | | 1) 0.956<br>2) 0.760<br>3) 0.856<br>4) 0.834 | 1) 0.915<br>2) 0.747<br>3) 0.844<br>4) 0.802 | 1) 0.985<br>2) 0.966<br>3) 0.960<br>4) 0.956 | |
| 2023 | 3 | pLM4Alg | Frozen pre-trained embeddings from ESM2 | Aggregated by encoding (Average pooling) | ALLERGOME, COMPARE, AllergenOnline, WHO/IUIS, UniProt | UniProt | 8029 | 8029 | NA | NA | Allergens and non-allergens share the same length distribution | 20% of the dataset was randomly sampled to form the test set | None | Deep learning model with convolutional layers | 0.951 | 0.942 | 0.96 | | | 0.902 | 0.99 | |
| | 4 | NetAllergen | MHC-II propensity predicted by NetMHCIIpan 4.0, physicochemical properties, structural features predicted by NetSurfP 3.0 | Aggregated by encoding | AllergenOnline | NCBI BLAST database | 537 | 2477 | For the training set, the sequence similarity between each allergen in the dataset was determined by BLASTP using an all-against-all alignment. If the query did not provide a significant hit to a protein in the database, the E-value was set to a value of 10. Using the pairwise comparison matrix of E-values, allergens with low E-values were clustered together using Minimum Spanning Tree (MST) algorithm. All clusters were then randomly split into training and test sets for 5-fold CV | BLASTP hits of each allergen from the same species and with an E-value higher than 1E-10 were retained ("… to avoid including sequences with high similarity to known allergens that could be untested antigens …") | Only non-allergens with <20% difference in length were retained for each allergen | 1) A dataset was compiled with 765 allergens and 2015 non-allergens from the AlgPred2 evaluation set, after removing identical sequences<br>2) A dataset was compiled using 1065 allergens and 4914 non-allergens from the WHO/IUIS, Allerbase, COMPARE, Allergome, and SDAP databases, after removing identical sequences | None | RF | | | | 1) 0.712 | | | 1) 0.886 | |
| | 5 | Kumar et. al., 2023 | Physicochemical properties, molecular properties, structural descriptors | NA | WHOIUIS, SDAP, AllergenOnline | PDB, UniProt, Swiss-Prot database, SCOP | 2427 | 2427 | NA | NA | NA | 30% of the dataset was randomly sampled to form the test set | None | Ensemble method with Extra Tree Classifier, DBN Classifier, and Cat Boost | 0.892 | | 0.892 | | 0.89 | 0.783 | 0.892 | 0.921 (0.512) |
| | 6 | allerStat | K-mer counts | k-mers | COMPARE | UniProt reviewed proteins of 20 food species | 2248 | 18906 | NA | NA | NA | None | Leave-category-out cross-validation, where each category corresponds to a biological species. | SVM | | | | | 0.517 | 0.45 | 0.873 | |
| | 7 | DeepAlgPro | One-hot encoding | NA | SDAP, WHO/IUIS, AllergenOnline, COMPARE, UniProt, AllerTop v.2 dataset, AlgPred 2.0 dataset | UniProt | 3550 | 3550 | NA | Non-allergens with a mutual sequence similarity of over 40% and a similarity to allergens greater than 40%, as calculated by CD-HIT, were removed | NA | 1) 710 allergens and non-allergens were randomly sampled from the training set | 2) 24 novel allergens after a certain timepoint from the IUIS Allergen Nomenclature but were not included in the training and test dataset were selected to form the external test set | Deep learning model with convolutional and attention layers | 1) 0.916<br>2) 0.875 | 1) 0.903<br>2) 1.0 | | 1) 0.928<br>2) 1.0 | 1) 0.915<br>2) 0.933 | | 1) 0.963 | 1) 0.969 (0.5) |
| | 8 | Nedyalkova et.al., 2023 | 74 physicochemical properties | Aggregated by encoding | CSL, FAARP, SDAP | Swiss-Prot | 954 | 954 | NA | NA | NA | 5-fold CV | None | K-NN (k=1) | 0.93 | 0.9 | 0.95 | | | 0.86 | 0.98 | |
| 2022 | 9 | ProAll-D | E-descriptors, ACC | Aggregated by encoding | CSL | NCBI | 2427 | 2427 | NA | NA | NA | 20% of the dataset was randomly sampled to form the test set | None | LSTM | 0.915 | 0.91 | | 0.91 | 0.91 | | 0.915 | |
| | 10 | AllerCatPro 2.0 | Amino-acid sequence | NA | WHO/IUIS, UniProtKB, Allergome, COMPARE, FARRP | NA | 4979 | NA | NA | Non-allergens with the same fold as allergens were selected for the external test set | NA | 1) The dataset contained 2003 allergens and 2015 non-allergens from the AlgPred2 validation set | 2) The dataset contained 218 allergens structurally non-redundant with other allergens, as calculated by CLICK, and 212 non-allergens with the same structure fold as the allergens<br>3) 20 allergens that were added to the WHO/IUIS Allergen Nomenclature database between December 2021 and April 2022 | Decision workflow (features include presence of gluten-like repeats, presence of epitopes, sequence identity, k-mer counts) | 1) 0.960<br>2) 0.847<br>3) 0.950 | 1) 0.932<br>2) 1.000<br>3) 0.970 | 1) 0.988<br>2) 0.689 | | | 1) 0.921<br>2) 0.727 | | |
| 2021 | 11 | AlgPred 2.0 | AAC | Aggregated by encoding | COMPARE, AllergenOnline, AlgPred dataset, AllerTop dataset, Swiss-Prot | Swiss-Prot, AlgPred dataset, AllerTop dataset | 10075 | 10075 | Training and test sets share <40% identity by CD-HIT | NA | NA | 1) 20% of the dataset was randomly sampled to form the test set | 2) The dataset included 297 allergens sourced from the COMPARE database after its 2020 release and from Swiss-Prot version 2020_04<br>3) Dataset 2) was filtered using CD-HIT to remove sequences with 40% or greater sequence identity to the internal dataset | Ensemble method (RF, MERCI, BLAST) | 1) 0.942<br>2) 0.942<br>3) 0.911 | 1) 0.931<br>2) 0.942<br>3) 0.911 | 1) 0.954 | | | 1) 0.88 | 1) 0.99 | |
| | 12 | Wang et al., 2021 | One-hot encoding | Truncated to fixed length (length not specified) | WHO/IUIS, SDAP, NCBI | NCBI | 583 | 600 | NA | NA | NA | 20% of the dataset was randomly sampled to form the test set | None | Frozen ProtBERT encoder followed by a FFNN prediction head | 0.931 | 0.942 | | 0.926 | | | | |
| 2019 | 13 | AllerCatPro | Amino-acid sequence | NA | WHO/IUIS, UniProtKB, Allergome, COMPARE, FARRP | NA | 4180 | NA | NA | Non-allergens with the same fold as allergens were selected for the external test set | NA | None | The dataset contained 221 allergens structurally non-redundant with other allergens, as calculated by CLICK, and 221 non-allergens with the same fold as the allergens | Decision workflow (features include presence of gluten-like repeats, presence of epitopes, sequence identity, k-mer counts) | 0.84 | 1 | 0.67 | | | | | |
| 2014 | 14 | AllerTOP v.2 | E-descriptors, ACC | Aggregated by encoding | CSL, FARRP, AllergenOnline, SDAP, Allergome | Swiss-Prot | 2427 | 2427 | NA | NA | NA | 5-fold CV | None | K-NN (k=1) | 0.887 | 0.867 | 0.907 | 0.903 | 0.885 | 0.775 | | |
| | 15 | Allerdictor | K-mer counts (k=6) | Aggregated by encoding | WHO/IUIS, Allergome, SDAP, AllergenOnline, AllerMatch | Swiss-Prot | Dataset A) 3907<br>Dataset B) 1990<br>Dataset C) 1662 | Dataset A) 39070<br>Dataset B) 19900<br>Dataset C) 16620 | Dataset A) NA<br>Dataset B) No sequence in the training set shares 95% or more identity with any sequence in the test set, as determined by BLASTClust<br>Dataset C) No sequence in the training set shares 50% or more identity with any sequence in the test set, as determined by BLASTClust | NA | NA | 1) 10-fold CV of Dataset A<br>2) 10-fold CV of Dataset B<br>3) 10-fold CV of Dataset C<br>4) Allerdictor was trained on Dataset A and tested on the entire Swiss-Prot with 539616 sequences<br>5) Allerdictor was trained on Dataset B and tested on the entire Swiss-Prot with 539616 sequences<br>6) Allerdictor was trained on Dataset C and tested on the entire Swiss-Prot with 539616 sequences<br>7) 167 allergens and 1663 non-allergens randomly drawn from dataset C | None | SVM | 7) 0.94 | 7) 0.56 | | 4) 0.354<br>5) 0.267<br>6) 0.454<br>7) 0.68 | | 7) 0.58 | | 1) 0.97 (0.091)<br>2) 0.91 (0.091)<br>3) 0.85 (0.091)<br>7) 0.59 (0.091) |
| | 16 | AllergenFP | E-descriptors, ACC | Aggregated by encoding | CSL, FARRP, AllergenOnline, SDAP, Allergome | Swiss-Prot | 2427 | 2427 | NA | NA | NA | 1) Leave-one-out cross-validation<br>2) 10-fold CV | None | K-NN (k=1), distance by Tanimoto coefficient similarity | 1) 0.879<br>2) 0.85 | 1) 0.868<br>2) 0.85 | 1) 0.891<br>2) 0.85 | 1) 0.889<br>2) 0.85 | 1) 0.878<br>2) 0.85 | 1) 0.759<br>2) 0.70 | | |
| 2013 | 17 | AllerTOP | z-descriptors, ACC | Aggregated by encoding | CSL, FARRP, SDAP | NCBI | 2210 | 2210 | NA | Non-allergens were proteins with no BLAST matches with any allergens at E-value < 0.001 | NA | 5-fold CV | None | K-NN (k=3) | 0.879 | 0.868 | 0.78 | 0.799 | 0.836 | 0.671 | | |

**Supplementary Table 1 Summary of published allergen prediction methods.** This table provides a detailed comparison of 17 allergen prediction tools published over the last decade. The comparison covers key aspects of their implementation, including feature encoding, dataset sources and composition, controls for inter-split (training vs. test) and inter-class (allergen vs. non-allergen) sequence similarity, evaluation strategies, and reported performance metrics. The "External test sets" column specifies the type of external validation used (if any), often based on criteria such as newly added sequences (By Date) or sequences from species not present in the training data. For the rule-based AllerCatPro and AllerCatPro 2.0, which lack a formal training process, their specific test sets composed of structurally non-redundant proteins are also included in this category. Abbreviations: AAC: Amino-Acid Composition; AAC-PSSM: Amino-Acid Composition - Position-Specific Scoring Matrix; ABT: Adaptive Boosting; ACC: Auto-Cross Covariance; CB: CatBoost; CLICK: A computational algorithm for protein structure comparison; CNN: Convolutional Neural Network; DBN: Deep Belief Network; DDE: Dipeptide Deviation from Expected Mean; DNN: Deep Neural Network; FFNN: Feed-Forward Neural Network; GBT: Gradient Boosting Trees; K-NN: K-Nearest Neighbors; LGBT: Light Gradient Boosting Machine; LSTM: Long Short-Term Memory; MERCI: Motif-EmeRging and with Classes-Identification; PseAAC: Pseudo Amino-Acid Composition; RF: Random Forest; SVM: Support Vector Machine; XGBT: Extreme Gradient Boosting.

| WHOIUIS Nomenclature | Wild-type or Mutated | Length | Allergenicity | Mutation | Reference |
|---|---|---|---|---|---|
| Api g 1.0101 | Wild-type | 154 | 1 | None | 79 |
| Api g 1.0101 | Mutated | 154 | 0 | S112P | 79 |
| Api g 1.0201 | Wild-type | 159 | 1 | None | 79 |
| Api g 1.0201 | Mutated | 159 | 1 | C112P | 79 |
| Api m 1 | Wild-type | 134 | 1 | None | 80 |
| Api m 1 | Mutated | 124 | 0 | I1S, Y3S, H11A, K14E, S16A, N19L, E20K, R23E, K25E, H26A, A29K, R31Q, T32A, H33Q, M35S, S46R, T51E, T53R, S55A, D62E, K66A, D69E, N73K, C105P, del106-115 | 80 |
| Cra g 1 | Wild-type | 284 | 1 | None | 91 |
| Cra g 1 | Mutated | 195 | 0 | del44-49, del69-85, del94-109, del134-144, del209-224 | 91 |
| Cra g 1 | Mutated | 284 | 0 | T44V, S45A, T69E, E73L, T77K, S79T, E80D, Q83G, E84D, I85V, M99L, E100D, S102A, E103Q, Q107A, N134A, N135M, A136K, S137D, R140K, T141M, D142E, V143L, L144Q, V209A, N211A, D212E, Q213K, A214Y, R217K, S220K, T224E | 91 |
| Cra g 1 | Mutated | 284 | 0 | T44Q, S45A, T69E, E73Q, T77K, S79T, E80D, Q83A, E84D, I85V, M99L, E100D, S102A, E103Q, Q107A, N134A, N135M, A136K, S137D, R140K, T141M, D142E, V143L, L144Q, V209A, Q210S, N211E, D212E, Q213E, A214Y, Q216T, R217K, S220K, T224E | 91 |
| Cra g 1 | Mutated | 284 | 0 | T44V, S45A, T69E, T77K, E100D, S102A, E103Q, Q107A, T141M, L144Q, V209A, N211A, Q213K | 91 |
| Der f 2 | Wild-type | 129 | 1 | None | 81 |
| Der f 2 | Mutated | 129 | 1 | C21S, C27S | 81 |
| Der f 2 | Mutated | 129 | 1 | C21S | 81 |
| Der f 2 | Mutated | 129 | 1 | C27S | 81 |
| Der f 2 | Mutated | 129 | 0 | C8S | 81 |
| Der f 2 | Mutated | 129 | 0 | P34A, P35A, P99A | 82 |
| Der f 2 | Mutated | 129 | 0 | C119S | 81 |
| Der f 2 | Mutated | 129 | 0 | C73S, C78S | 81 |
| Der f 2 | Mutated | 129 | 0 | C73S | 81 |
| Der f 2 | Mutated | 129 | 0 | C78S | 81 |
| Der f 2 | Mutated | 129 | 0 | C8S, C119S | 81 |
| Der p 21 | Wild-type | 121 | 1 | None | 83 |
| Der p 21 | Mutated | 121 | 0 | D82P, K110G, E77G, E87S | 83 |
| Der p 21 | Mutated | 121 | 0 | D82P, K110G | 83 |
| Equ c 1 | Wild-type | 172 | 1 | None | 84 |
| Equ c 1 | Mutated | 172 | 0 | V47K, V110E, F112K | 84 |
| Equ c 1 | Mutated | 172 | 0 | E21Y, V110E, F112K | 84 |
| Hev b 5 | Wild-type | 151 | 1 | None | 85 |
| Hev b 5 | Mutated | 151 | 0 | T53A, E91A | 85 |
| Hev b 5 | Mutated | 151 | 0 | T53A, E91A, E108A | 85 |
| Hev b 5 | Mutated | 151 | 0 | E31A, T53A, E91A, E108A | 85 |
| Hev b 5 | Mutated | 151 | 0 | T18A, E31A, T53A, E91A, E108A | 85 |
| Hev b 5 | Mutated | 151 | 0 | T18A, E31A, T53A, T78A, E91A, E108A | 85 |
| Hev b 5 | Mutated | 151 | 0 | T18A, E31A, T53A, T78A, E91A, E108A, T134A | 85 |
| Hev b 5 | Mutated | 151 | 0 | T18A, E31A, E50A, T53A, T78A, Q87A, E91A, E108A, K110A, T134A | 85 |
| Hev b 5 | Mutated | 151 | 0 | T53A, E91A, K93A, P94A, E108A | 85 |
| Hev b 5 | Mutated | 151 | 0 | T53A, E91A, K93A, P94A, E108A, K125A, P126A | 85 |
| Hev b 5 | Mutated | 151 | 0 | T53A, E70A, E91A, K93A, P94A, E108A, K125A, P126A | 85 |
| Hev b 5 | Mutated | 151 | 0 | T18A, E31A, E50A, T53A, T78A, Q87A, E91A, K93A, P94A, E108A, K110A, T134A | 85 |
| Hev b 5 | Mutated | 151 | 0 | T18A, E31A, E50A, T53A, T78A, Q87A, E91A, K93A, P94A, E108A, K110A, K125A, P126A, T134A | 85 |
| Hev b 5 | Mutated | 151 | 0 | T18A, E31A, E50A, T53A, E70A, T78A, Q87A, E91A, K93A, P94A, E108A, K110A, K125A, P126A, T134A | 85 |
| Lol p 5 | Wild-type | 339 | 1 | None | 86 |
| Lol p 5 | Mutated | 339 | 1 | G336A, K338A | 86 |
| Lol p 5 | Mutated | 334 | 1 | del335-339 | 86 |
| Lol p 5 | Mutated | 339 | 1 | A247G, V248A, K249A | 86 |
| Lol p 5 | Mutated | 339 | 0 | K94A | 86 |
| Lol p 5 | Mutated | 339 | 0 | K215N, F216L, T217A, V218A | 86 |
| Lol p 5 | Mutated | 339 | 0 | K94A, K215N, F216L, T217A, V218A, G336A, K338A | 86 |
| Lol p 5 | Mutated | 339 | 0 | K94A, A247G, V248A, K249A, G336A, K338A | 86 |
| Lol p 5 | Mutated | 334 | 0 | K94A, K215N, F216L, T217A, V218A, del335-339 | 86 |
| Lol p 5 | Mutated | 334 | 0 | K94A, A247G, V248A, K249A, del335-339 | 86 |
| Mal d 1.0108 | Wild-type | 159 | 1 | None | 87 |
| Mal d 1.0108 | Mutated | 159 | 0 | T11P, I31V, T58N, T113C, I114V | 87 |
| Pen a 1 | Wild-type | 284 | 1 | None | 88 |
| Pen a 1 | Mutated | 284 | 0 | L46M, L53T, L95V, S136K, E145D, V190S, V199N, R255D, L260V, S269F, T277A, F278L | 88 |
| Phl p 5 | Wild-type | 284 | 1 | None | 89 |
| Phl p 5 | Mutated | 279 | 0 | del57P, del58P, del117P, del180P, P211L, del229P | 89 |
| Phl p 5 | Mutated | 278 | 0 | del57P, del58P, del117P, del146P, del180P, P211L, del229P | 89 |
| Phl p 5 | Mutated | 278 | 0 | del57P, del58P, del117P, del155P, del180P, P211L, del229P | 89 |
| Pru av 1 | Wild-type | 160 | 1 | None | 90 |
| Pru av 1 | Mutated | 160 | 1 | G112A | 90 |
| Pru av 1 | Mutated | 155 | 1 | del155-159 | 90 |
| Pru av 1 | Mutated | 160 | 0 | E46W | 90 |
| Pru av 1 | Mutated | 160 | 0 | G112P | 90 |

**Supplementary Table 2 Wild-type and mutated allergens comprising the Mutations external test set.** This table lists the wild-type allergens and their corresponding mutated counterparts used for constructing the Mutations external test set. The “Mutation” column details the specific amino acid changes, including substitutions (e.g., S112P indicates a change from Serine to Proline at position 112) and deletions. The “Reference” column indicates the citation number corresponding to the source publication in the main reference list.

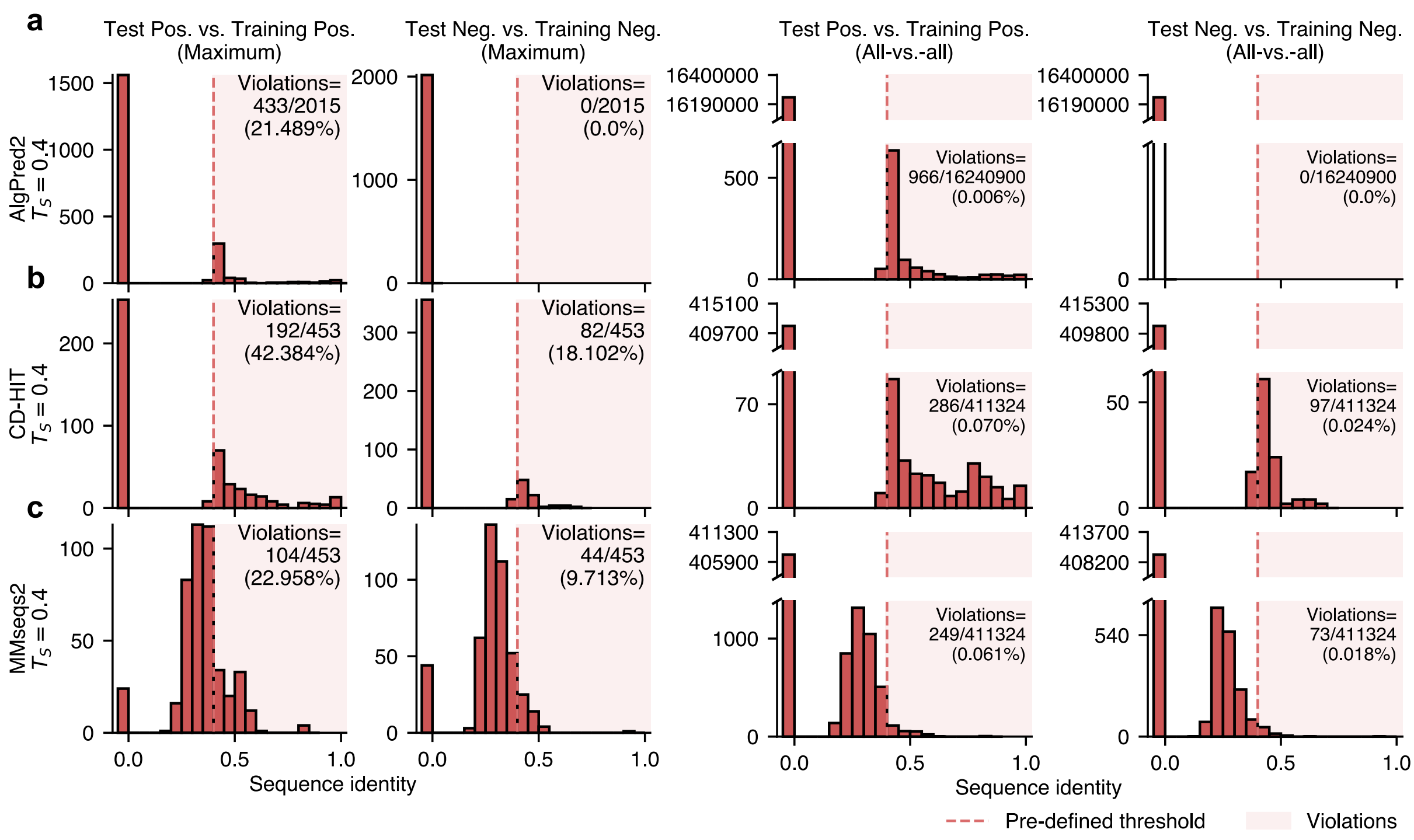

**Supplementary Fig. 1 Violations are present between training and test sets generated by CD-HIT and MMseqs2.** Distributions of sequence identity between test and training sets from: **a** the AlgPred2 dataset generated by CD-HIT, **b** our dataset generated by CD-HIT, and **c** our dataset generated by MMseqs2. "Violations" (highlighted region and text) denotes sequence pairs that exceed the pre-defined $T_S = 0.4$ identity threshold (dashed line), indicating potential data leakage between the test and training sets.

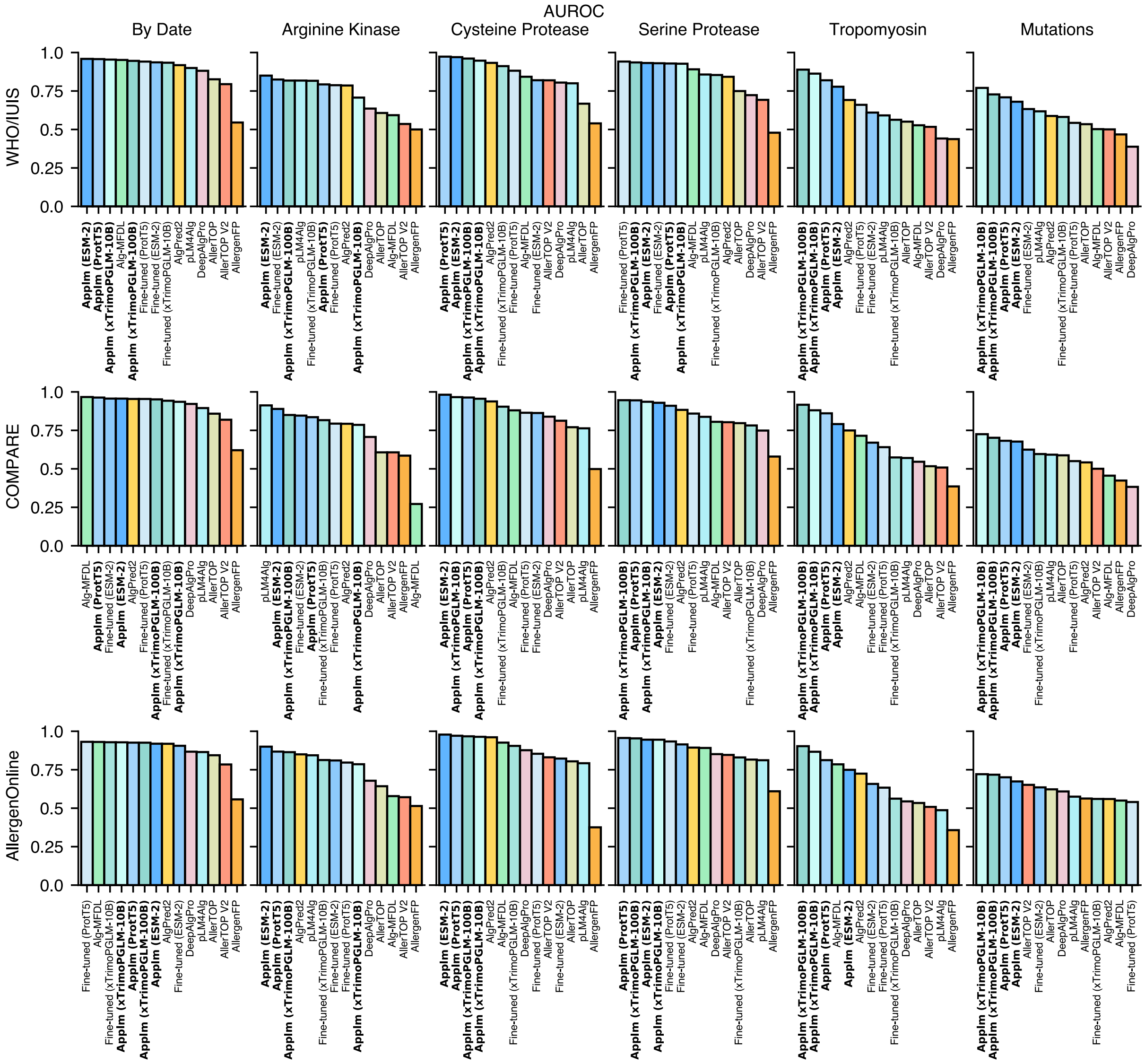

**Supplementary Fig. 2 Applm generally achieves top performance on the external benchmark based on AUROC.** The figure compares the AUROC of Applm models using different pLMs (bold) against LoRA fine-tuned pLMs and seven other methods. Each panel represents one of six external test sets (columns) for models trained on one of three datasets (rows).

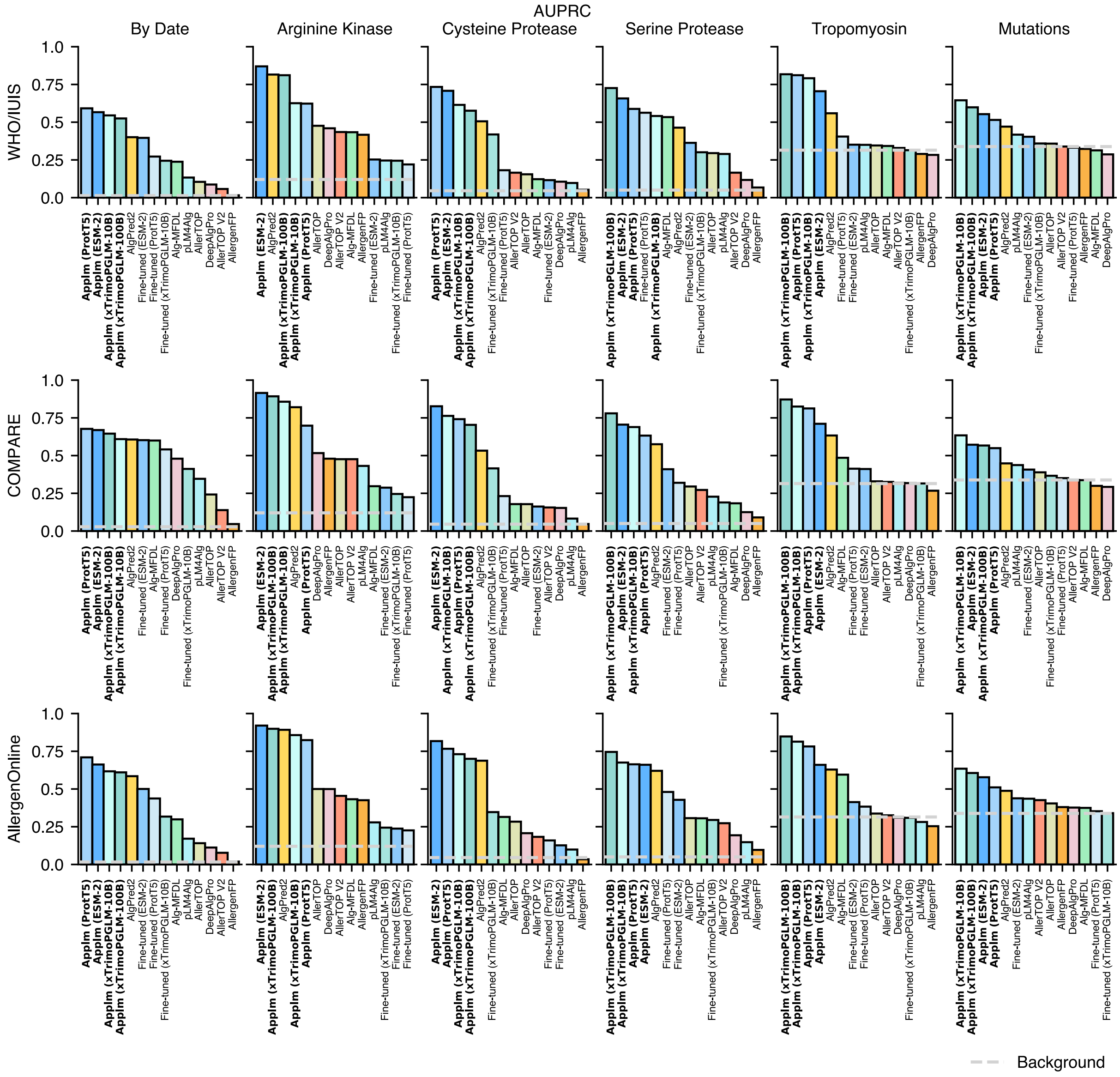

**Supplementary Fig. 3 Applm achieves top performance on the external benchmark based on AUPRC.** The figure compares the AUPRC of Applm models using different pLMs (bold) against LoRA fine-tuned pLMs and seven other methods. Each panel represents one of six external test sets (columns) for models trained on one of three datasets (rows).

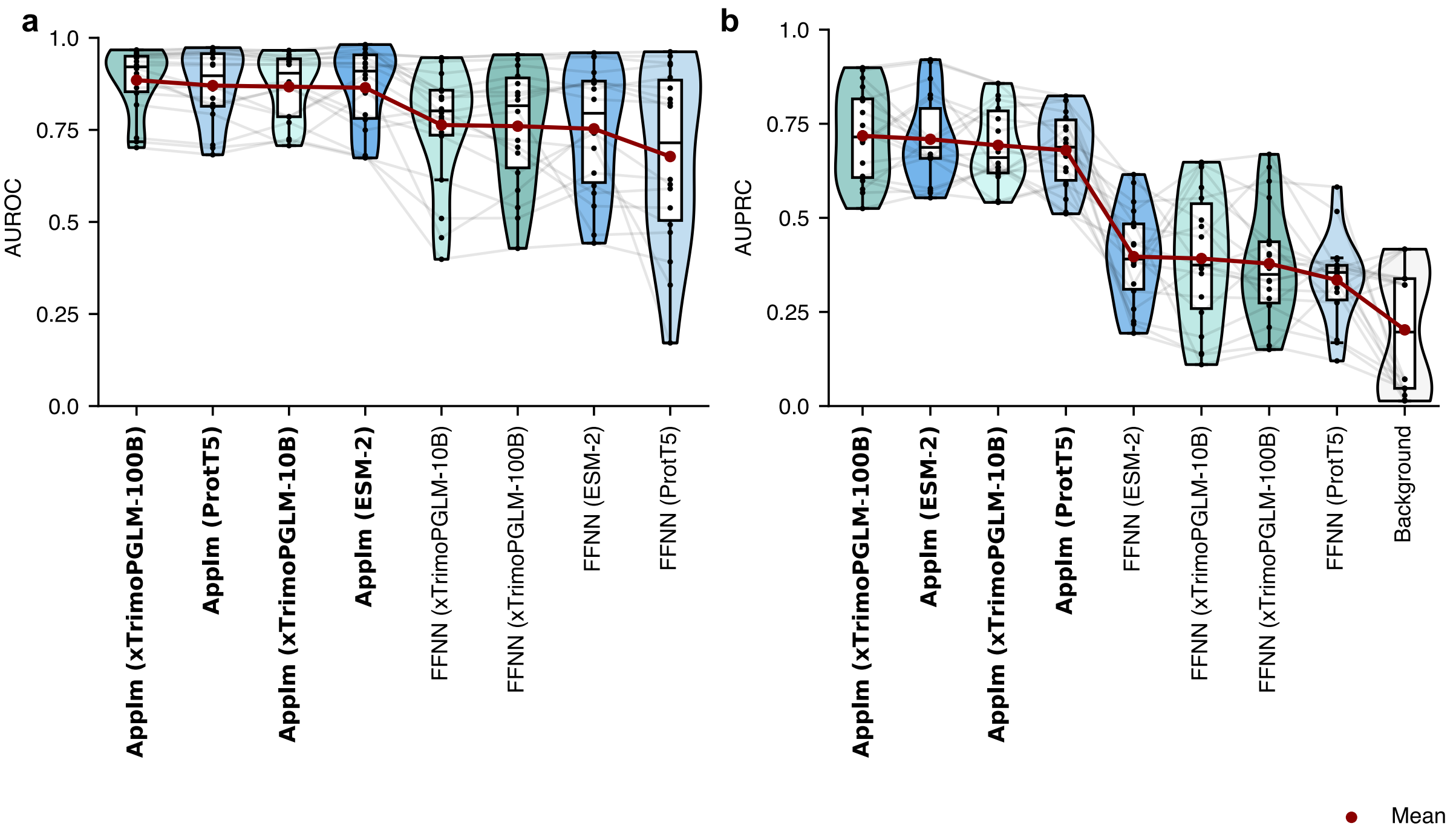

**Supplementary Fig. 4 Applm, our RF-based model, consistently outperforms an FFNN classifier using identical pLM embeddings.** The violin plots compare the performance distribution of Applm (bold) against FFNN for four different pLM embeddings. Performance is measured by **a** AUROC and **b** AUPRC across the six external test sets.

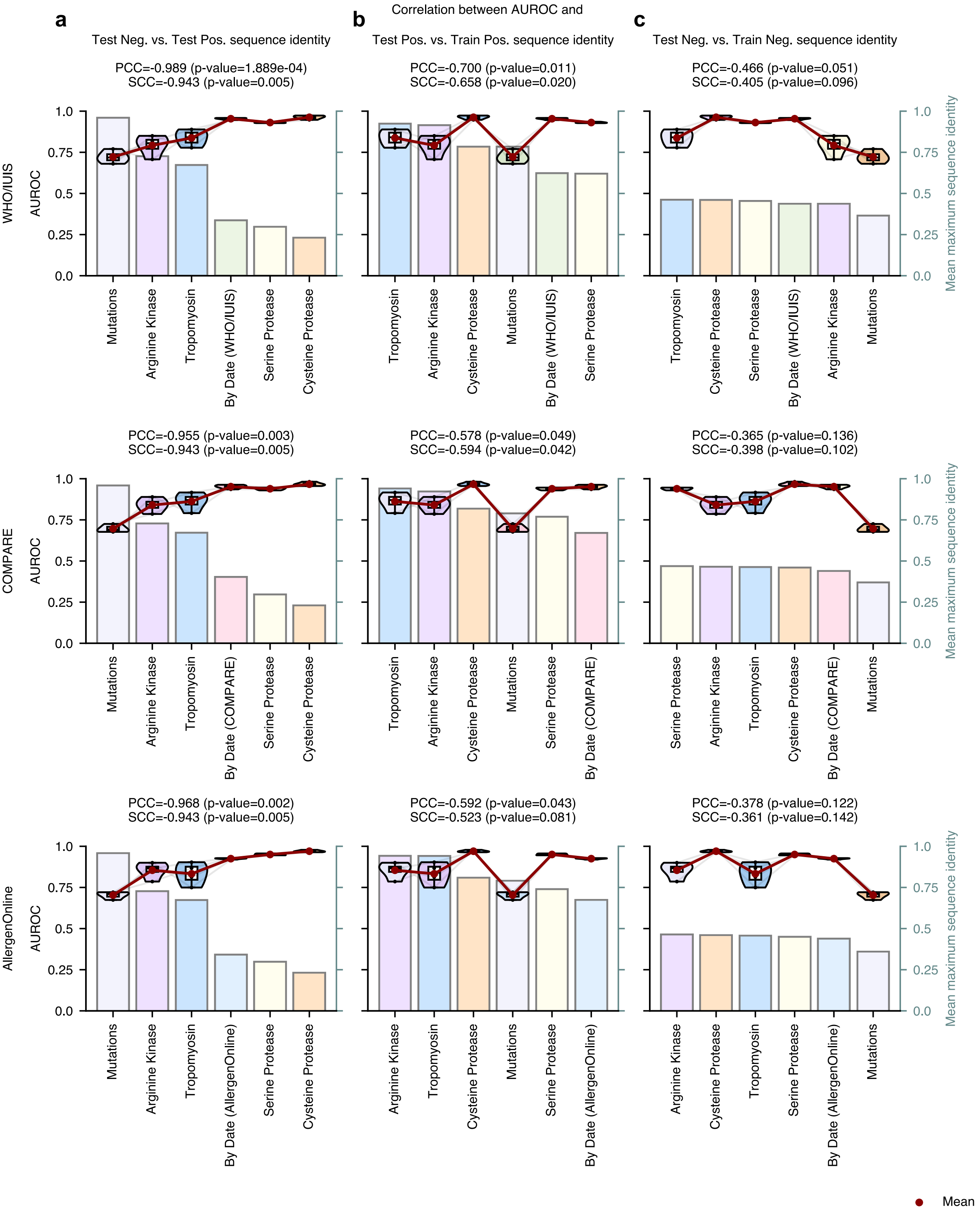

**Supplementary Fig. 5 Correlation between Applm's AUROC and sequence identity metrics.** Each panel correlates Applm's AUROC performance (violin plots, left y-axis) with a specific measure of mean maximum sequence identity (bars, right y-axis) across six external test sets. The rows correspond to models trained on the WHO/IUIS, COMPARE, or AllergenOnline datasets. The columns show correlations of average AUROC with the sequence identity between: **a** test negatives and test positives, **b** test positives and training positives, and **c** test negatives and training negatives. Pearson correlation coefficient (PCC) and Spearman's

rank correlation coefficient (SCC) values and their corresponding p-values are shown for each panel.

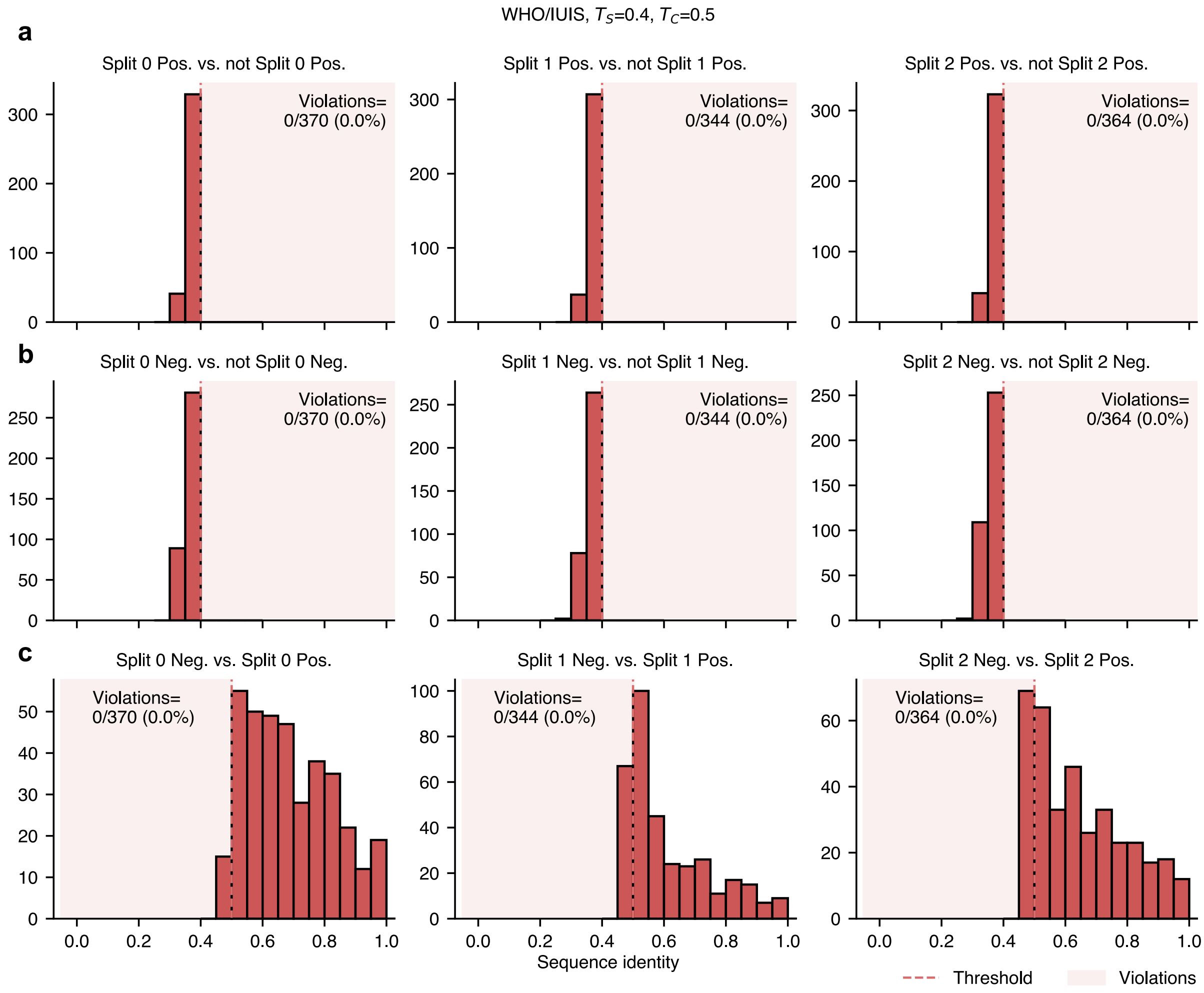

**Supplementary Fig. 6 Our similarity-aware pipeline does not produce violations.** Distributions of maximum sequence identity for a 3-fold CV split of the WHO/IUIS dataset, generated by our proposed pipeline: **a** inter-split identity between positive sequences, **b** inter-split identity between negative sequences, and **c** inter-class identity in each split. No violations are found in any comparison. The bar at the 0.5 mark in panel (**c**) is a visualization artifact from data binning of sequences with an identity exactly equal to the threshold, and does not represent a violation. We used $T_S = 0.4$ and $T_C = 0.5$ as an example. In this study, we carefully checked all settings across all $T_S$ and $T_C$ and confirmed no violations.

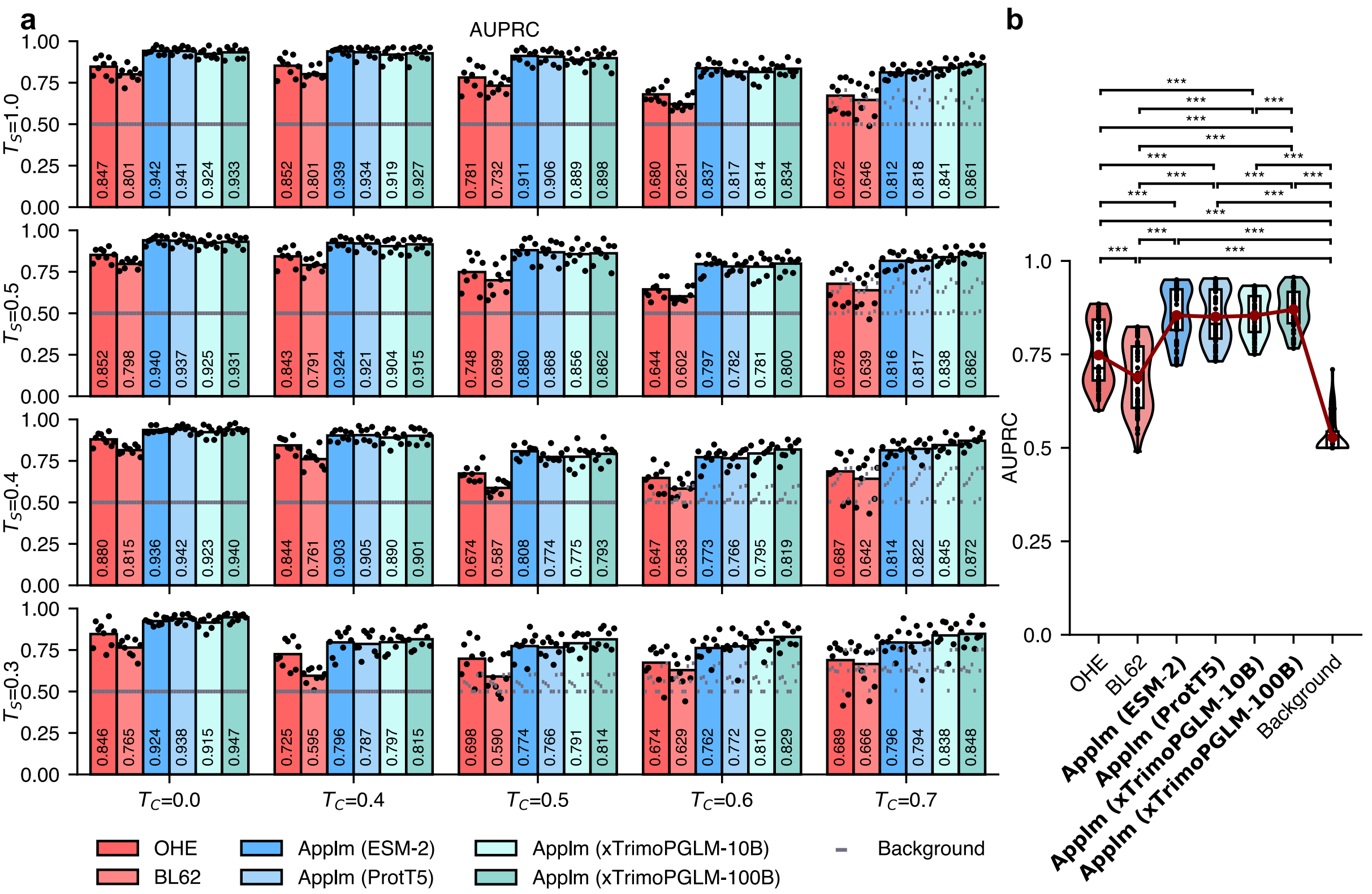

**Supplementary Fig. 7 Applm outperforms models using conventional encodings on similarity-aware internal CV. a** Bar plots showing the detailed AUPRC performance of Applm and models using conventional encodings across a grid of inter-split ($T_S$) and inter-class ($T_C$) similarity thresholds. Each dot represents an individual CV fold. **b** Violin plots comparing the overall AUPRC distributions of Applm against models using conventional encodings, aggregated from all conditions in (**a**).

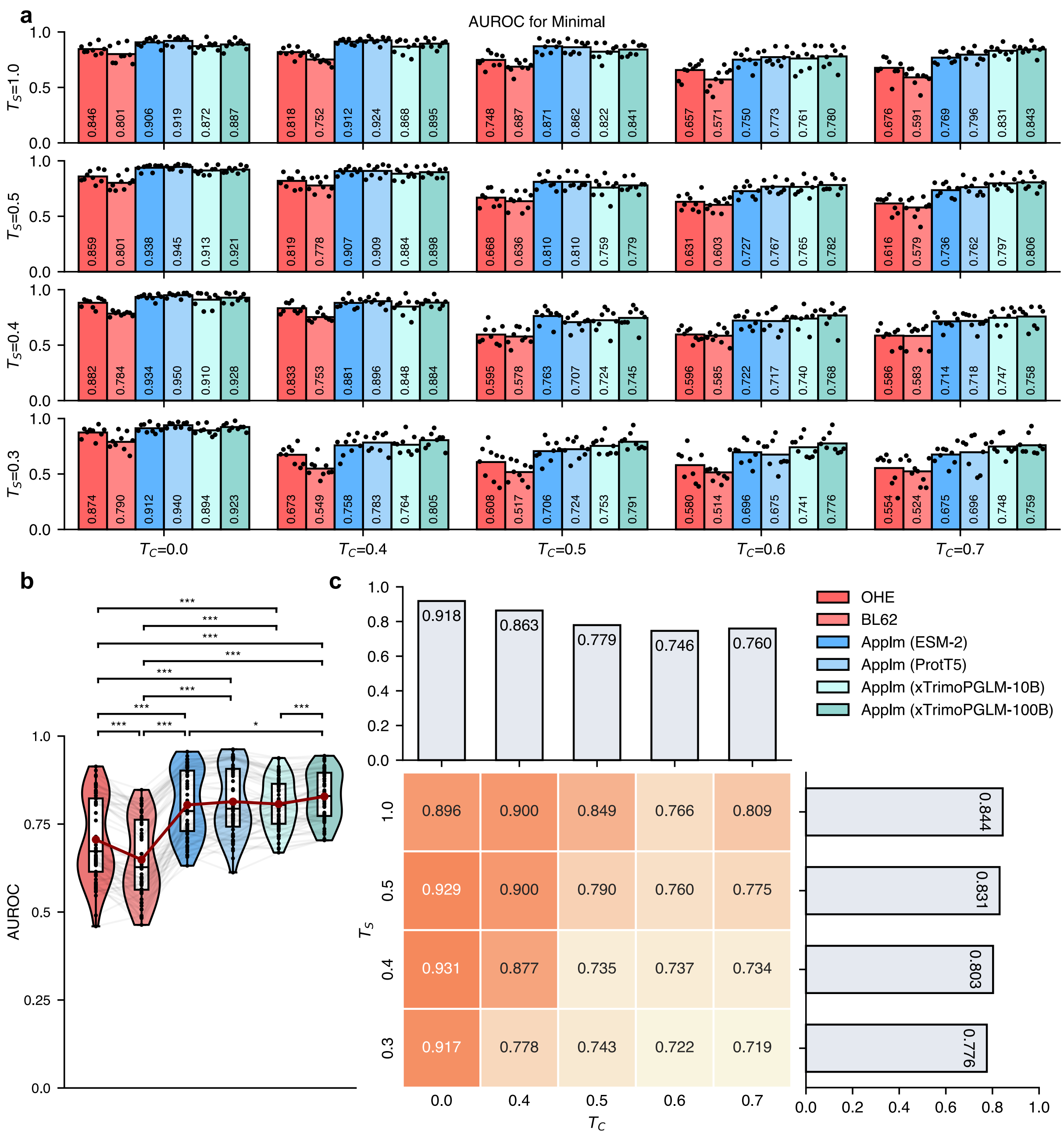

**Supplementary Fig. 8 Applm outperforms models using conventional encodings on similarity-aware internal CV under the Minimal strategy. a** Bar plots showing the detailed AUROC performance of Applm and models using conventional encodings across a grid of inter-split ($T_S$) and inter-class ($T_C$) similarity thresholds. Each dot represents an individual CV fold. **b** Violin plots comparing the overall AUROC distributions of Applm against models using conventional encodings, aggregated from all conditions in (**a**). **c** Heatmap of AUROC scores, averaged across Applm models leveraging different pLMs, illustrating the combined effect of $T_S$ and $T_C$. The bar plot at the top shows the performance for each $T_C$ level (averaged across all $T_S$ levels), while the bar plot on the right shows the performance for each $T_S$ level (averaged across all $T_C$ levels).

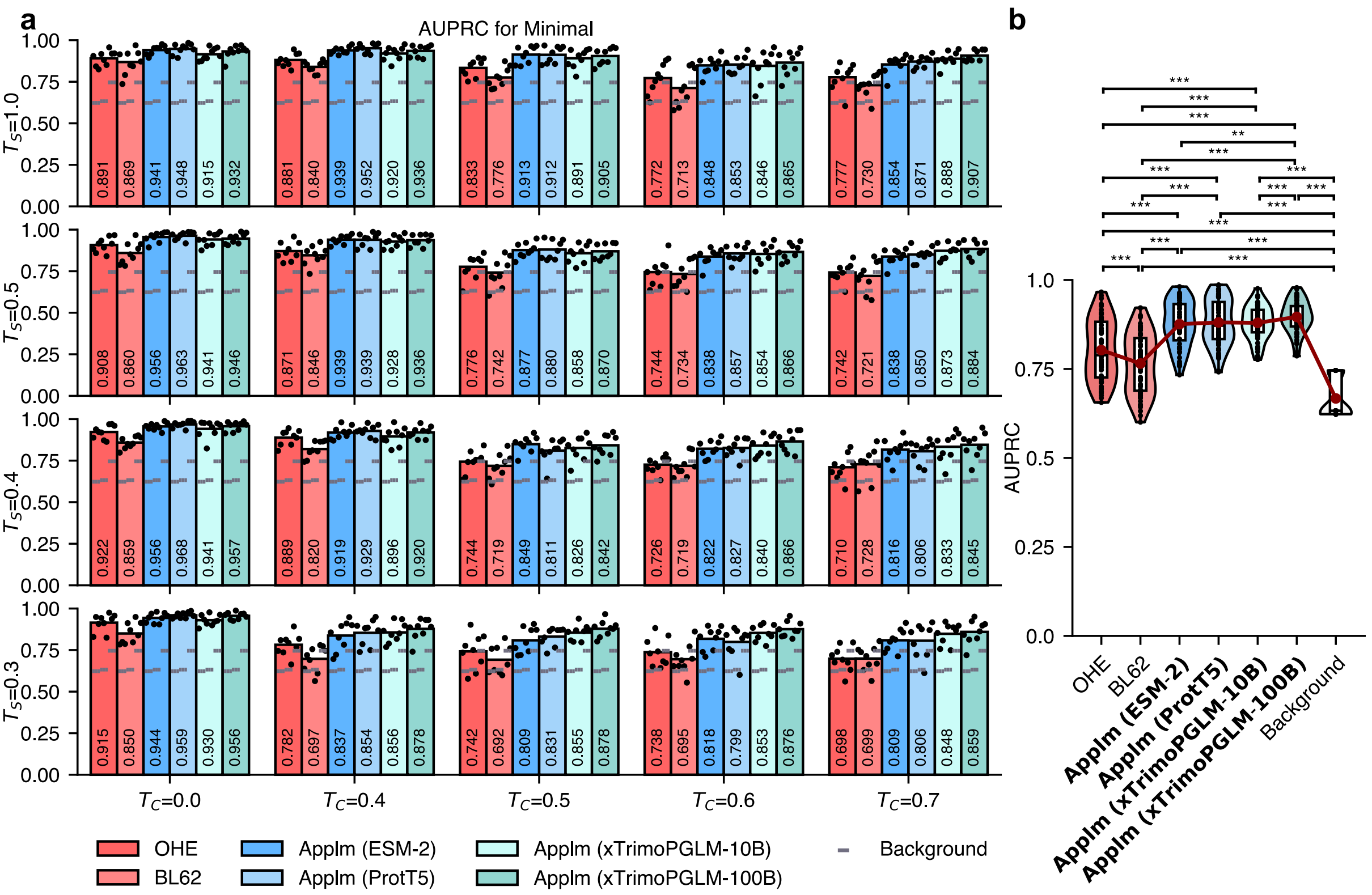

**Supplementary Fig. 9 Applm outperforms models using conventional encodings on similarity-aware internal CV under the Minimal strategy. a** Bar plots showing the detailed AUPRC performance of Applm and models using conventional encodings across a grid of inter-split ($T_S$) and inter-class ($T_C$) similarity thresholds. Each dot represents an individual CV fold. **b** Violin plots comparing the overall AUPRC distributions of Applm against models using conventional encodings, aggregated from all conditions in (**a**).

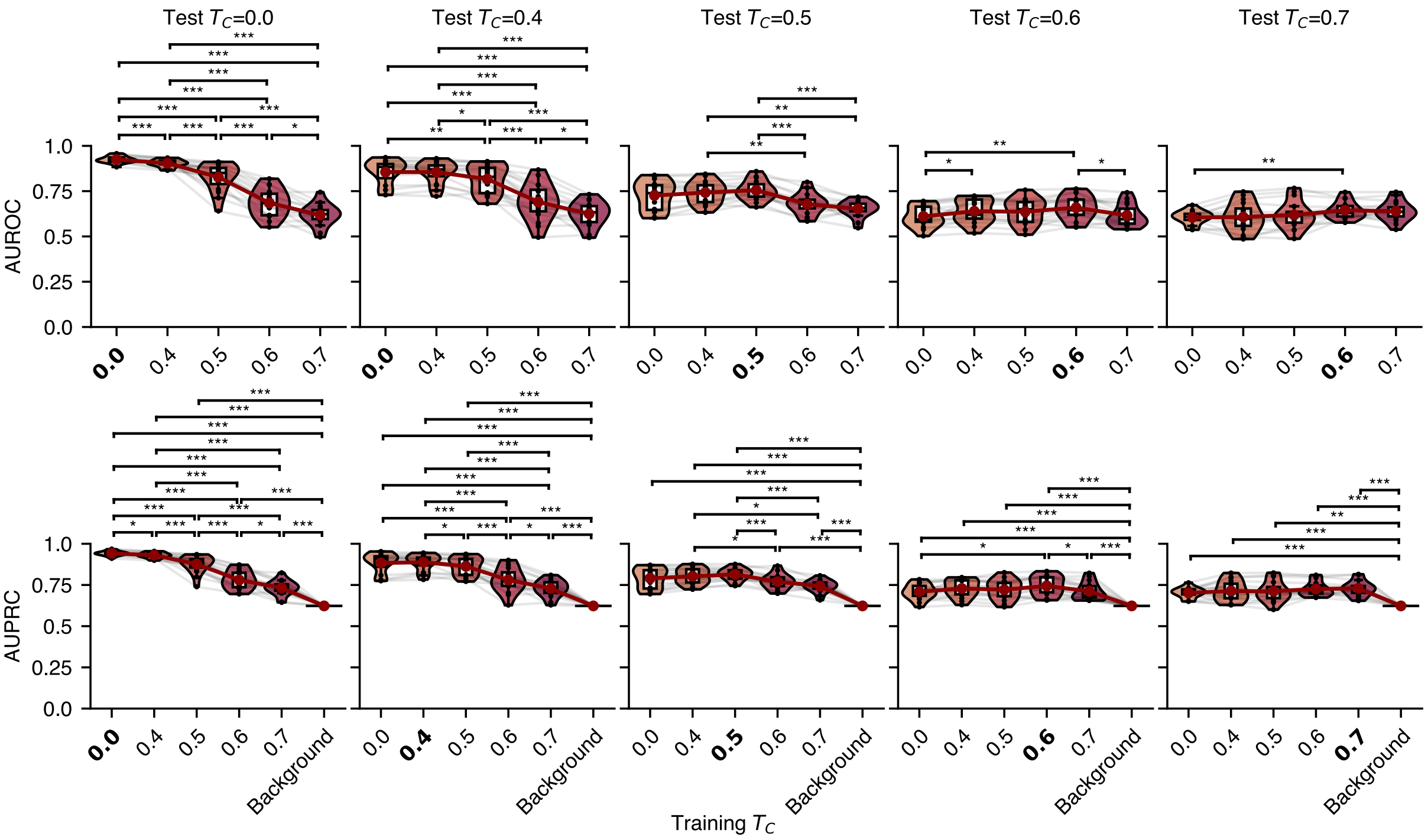

**Supplementary Fig. 10 Intrinsic difficulty impacts model performance under the Minimal strategy.** Performance grid showing AUROC and AUPRC where Applm was trained and tested on datasets with varying inter-class similarity thresholds ($T_C$). For each test $T_C$ (columns), the training $T_C$ (x-axis) that yields the best average performance is highlighted in bold.

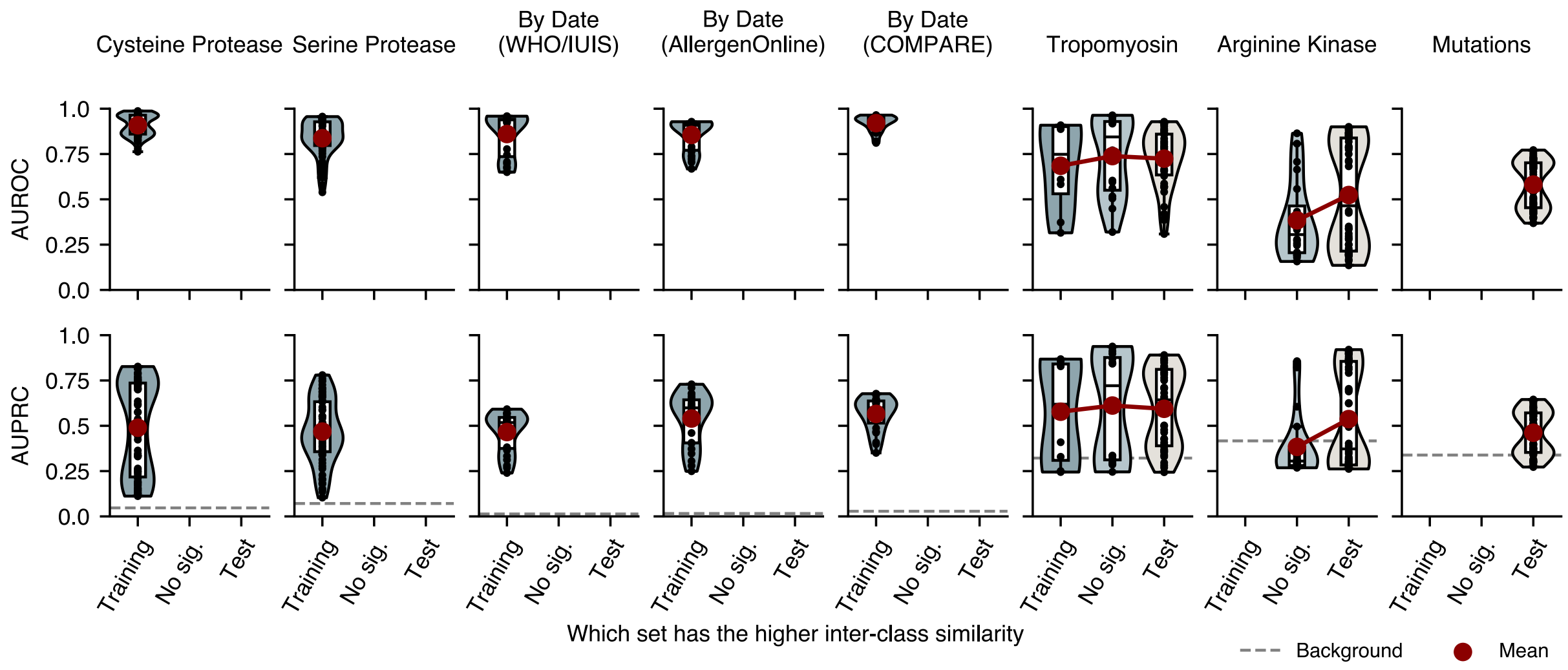

**Supplementary Fig. 11 Applm performance on external test sets, categorized by relative training and test set difficulty.** The figure displays Applm's AUROC (top) and AUPRC (bottom). Each panel corresponds to a specific training set-external test set pair, for which performance was evaluated under multiple experimental settings. Each setting was then categorized using a Mann-Whitney U test to compare the inter-class similarity (difficulty) between its training and test components. The categories are: "Training" (training set significantly harder), "Test" (test set significantly harder), or "No sig." (no significant difference) (Methods). Within each panel, the violin plots show the performance distribution of all settings for that specific pair, grouped by the resulting difficulty category.

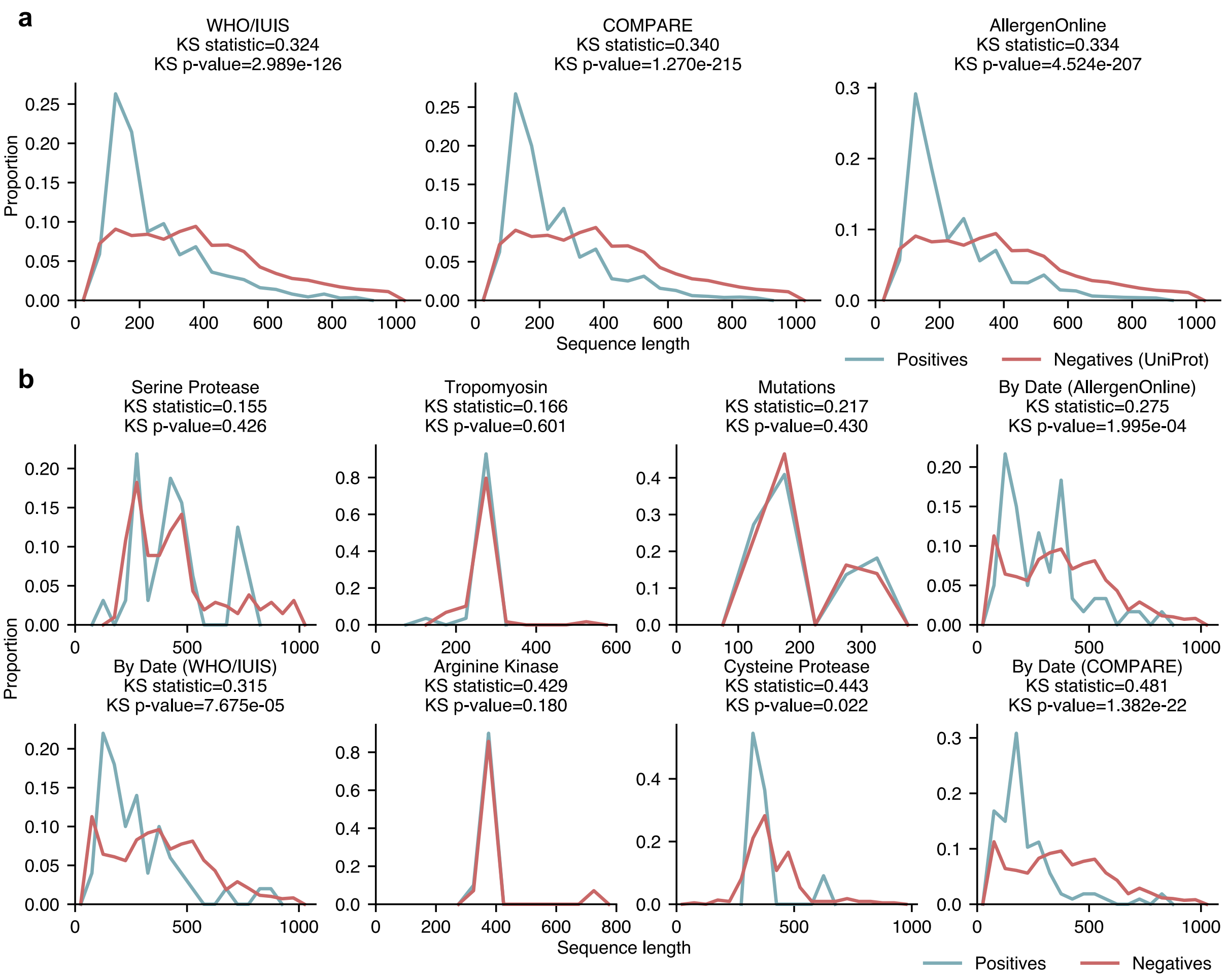

**Supplementary Fig. 12 Sequence length distributions for positive and negative sequences.** The distributions for each dataset were compared using a two-sample KS test, with the resulting KS statistic and p-value displayed in each panel. The KS statistic measures the distance between the two distributions; a small statistic combined with a large p-value indicates that the distributions are similar. **a** Distributions in the three training datasets. **b** Distributions in the external test sets.

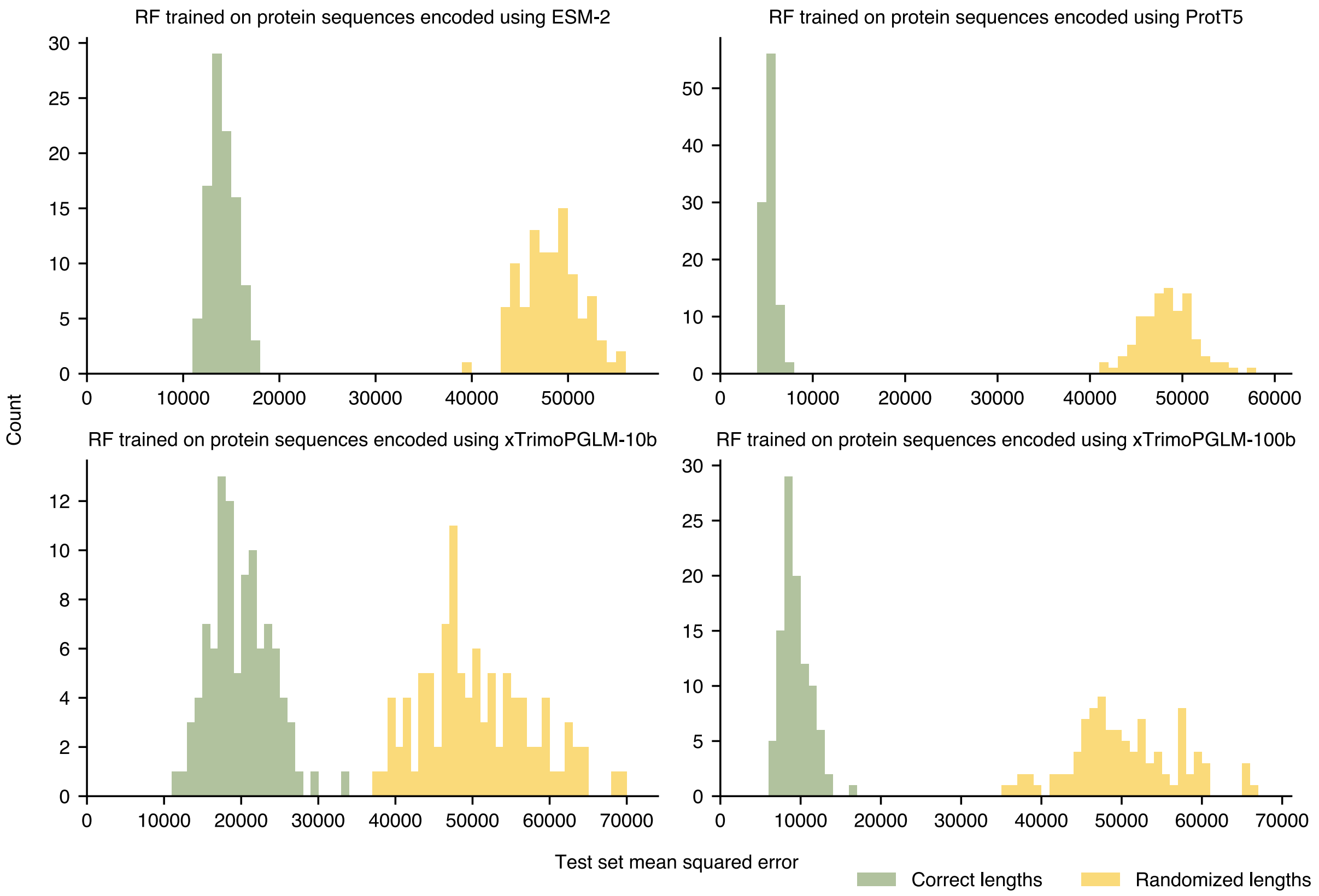

**Supplementary Fig. 13 pLM embeddings are highly predictive of sequence length.** To test if pLM embeddings contain information about sequence length, RF models were trained to predict this property from four different pLM embeddings using a dataset of 5,000 sequences. For each embedding type, the histograms show distributions of test set mean squared error (MSE) from 100 independent training runs. As a negative control, the entire procedure was repeated, but with the sequence lengths randomly shuffled across the 5,000 sequences before training. Each panel compares the MSE distributions from models trained on correct lengths (green) with those from the negative control (yellow).

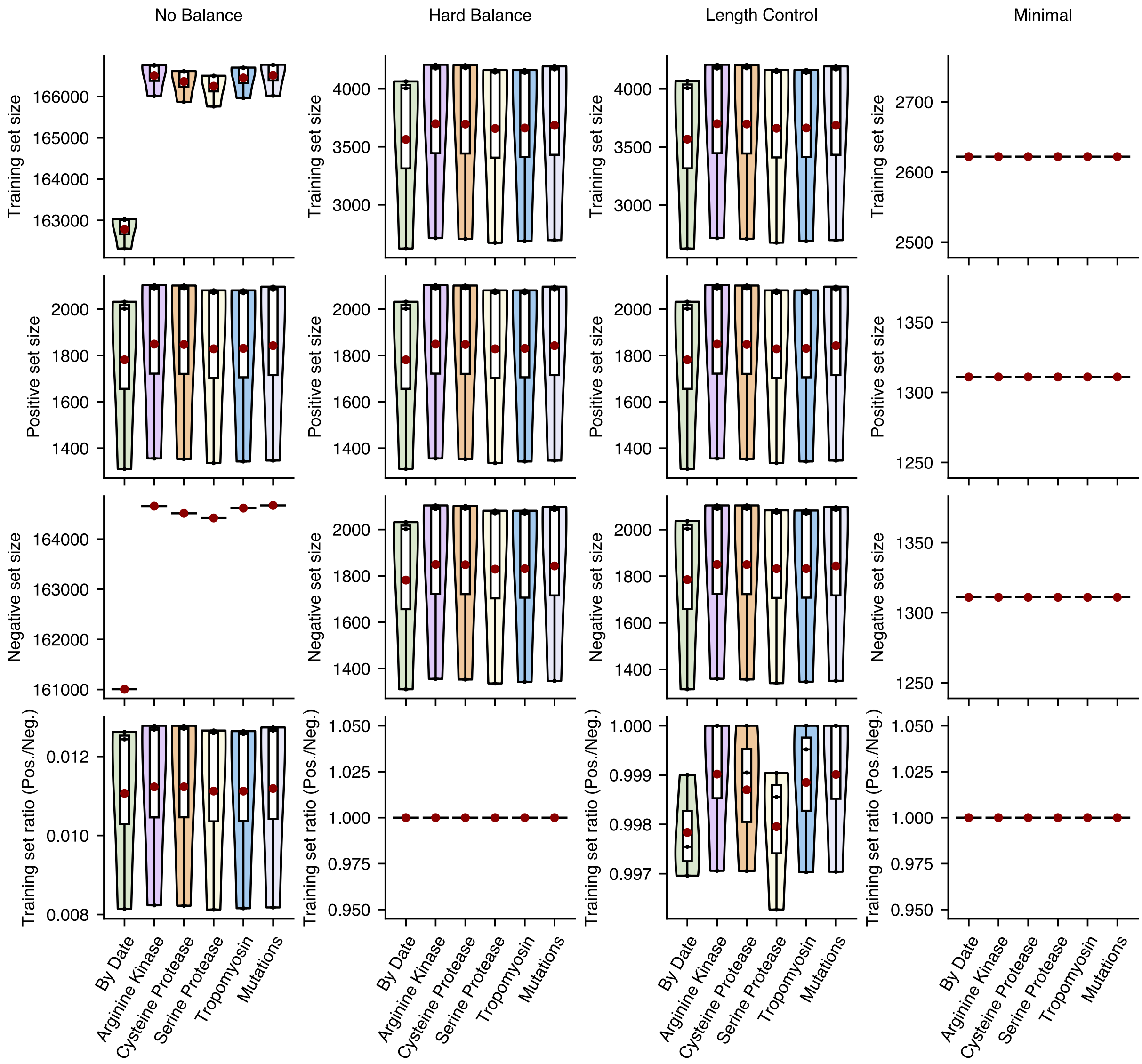

**Supplementary Fig. 14 Statistics of training sets for each external test set and construction strategy.** This figure shows statistics of training sets constructed using four strategies (columns) for various external test sets (x-axis). Each plot visualizes metrics from three distinct training sets, which were generated from the WHO/IUIS, COMPARE, and AllergenOnline databases.

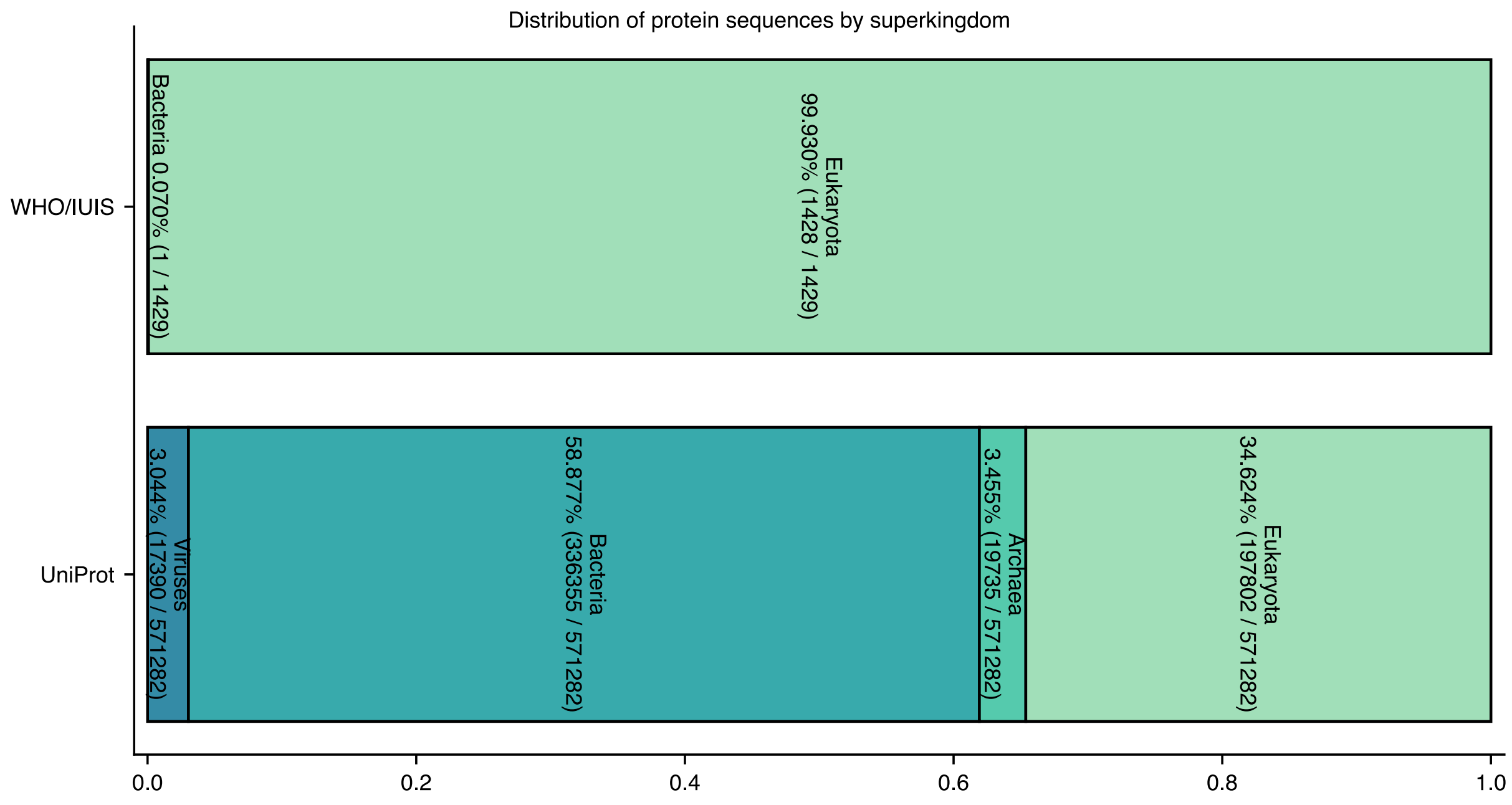

**Supplementary Fig. 15 Distribution of superkingdoms for protein sequences obtained from WHO/IUIS and UniProt.** The distribution of protein sequences by superkingdom was substantially different between allergen dataset WHO/IUIS and non-allergen dataset UniProt, which motivated us to only retain Eukaryotic sequences from UniProt.